\documentclass[preprint,12pt]{elsarticle}

\usepackage{amssymb}
\usepackage{amsmath}
\usepackage{bm}
\usepackage{subfigure}
\usepackage{adjustbox}
\usepackage{enumerate}
\usepackage{enumitem}
\newtheorem{mydef}{Definition}
\newtheorem{example}{Example}
\usepackage[linesnumbered, ruled, vlined]{algorithm2e}
\SetKwRepeat{Do}{do}{while}%
\usepackage{setspace}
\usepackage{booktabs}
\usepackage{multirow}

\usepackage{graphicx}
\usepackage{caption}
\usepackage[a4paper, left=2cm, right=2cm, top=2.5cm, bottom=2.5cm]{geometry}
\usepackage{float}

\journal{Nuclear Physics B}

\begin{document}

\begin{frontmatter}

\title{GAdaBoost: An Efficient and Robust AdaBoost Algorithm Based on Granular-Ball Structure} 

\author[inst1,inst2,inst3]{Qin Xie} 
\author[inst2,inst3]{Qinghua Zhang\corref{cor1}}
\author[inst2,inst3]{Shuyin Xia}
\author[inst1,inst2,inst3]{Xinran Zhou}
\author[inst2,inst4]{Guoyin Wang}

\cortext[cor1]{Qinghua Zhang. Email: zhangqh@cqupt.edu.cn}

\affiliation[inst1]{organization={School of Computer Science and Technology, Chongqing University of Posts and Telecommunications},
            city={Chongqing},
            postcode={400065},
            country={China}}

\affiliation[inst2]{organization={Chongqing Key Laboratory of Computational Intelligence, Chongqing University of Posts and Telecommunications},
            city={Chongqing},
            postcode={400065},
            country={China}}

\affiliation[inst3]{organization={Key Laboratory of Cyberspace Big Data Intelligent Security, Ministry of Education},
            city={Chongqing},
            postcode={400065},
            country={China}}

\affiliation[inst4]{organization={College of Computer and Information Science, Chongqing Normal University},
            city={Chongqing},
            postcode={401331},
            country={China}}

\begin{abstract}
Adaptive Boosting (AdaBoost) faces significant challenges posed by label noise, especially in multiclass classification tasks. Existing methods either lack mechanisms to handle label noise effectively or suffer from high computational costs due to redundant data usage. Inspired by granular computing, this paper proposes granular adaptive boosting (GAdaBoost), a novel two-stage framework comprising a data granulation stage and an adaptive boosting stage, to enhance efficiency and robustness under noisy conditions. To validate its feasibility, an extension of SAMME, termed GAdaBoost.SA, is proposed. Specifically, first, a granular-ball generation method is designed to compress data while preserving diversity and mitigating label noise. Second, the granular ball-based SAMME algorithm focuses on granular balls rather than individual samples, improving efficiency and reducing sensitivity to noise. Experimental results on some noisy datasets show that the proposed approach achieves superior robustness and efficiency compared with existing methods, demonstrating that this work effectively extends AdaBoost and SAMME.
\end{abstract}

\begin{keyword}
Granular Computing; Granular-Ball Structure; Adaptive Boosting Algorithm; Multiclass Classification

\end{keyword}

\end{frontmatter}

\section{Introduction}
\label{intro}
The basic idea behind ensemble learning is that for a complex classification task, integrating the judgments of multiple experts generally yields better results than relying on a single expert. The boosting algorithm \cite{56,9} is an ensemble learning technique. Boosting for classification combines several weak classifiers (e.g., tree stumps, which perform only slightly better than random guessing) in series, allowing them to learn from the data and correct errors in turn. Subsequent classifiers focus on samples misclassified by previous ones to improve performance.
Adaptive Boosting (AdaBoost) \cite{8,11} is the most advanced boosting algorithm. It adaptively adjusts sample weights, increasing the weight of misclassified samples at each iteration to ensure that subsequent base classifiers focus on these complex samples.

AdaBoost was initially developed for binary classification tasks. To address the broader applicability of real-world problems, which often involve multiclass classification scenarios, various extended algorithms have been proposed \cite{12,13,58}. Some extended AdaBoost algorithms reduce multiclass classification to multiple binary classifications, such as AdaBoost.MH \cite{14}. However, the time and memory required to build and store all the binary classifiers can be prohibitive for datasets with a large number of classes. Other extended AdaBoost algorithms directly utilize a multiclass classifier as the base learner to build a multiclass classification algorithm. Examples include AdaBoost.HM \cite{15}, AdaBoost.MM \cite{16}, and SAMME \cite{17}, etc., with SAMME being the most efficient and widely used multi-class AdaBoost algorithm.
For scenarios with a moderate dataset size, limited computational resources, and model interpretability requirements, AdaBoost is a good choice. Consequently, AdaBoost has also been applied to anomaly detection \cite{20}, image classification \cite{21}, unlabeled learning \cite{57}, and intelligent healthcare \cite{22}.

However, despite its many extensions, AdaBoost still suffers from the following limitations. First, AdaBoost is known to be sensitive to label noise and outliers because it prioritizes misclassified samples. Various methods have been proposed to address this limitation. For example, some works\cite{25,26,27,28} adopt the loss function modification strategy, and some works\cite{29,30,31,55} implement a weight update adjustment strategy. However, most existing studies have naturally focused on improving the robustness of AdaBoost in binary classification scenarios, as it was initially designed as a binary classification algorithm. In particular, \cite{32} proposed an extended multiclass AdaBoost algorithm based on SAMME, namely Rob\_SAMME, for handling mislabeled noisy data. However, Rob\_SAMME suffers from issues such as high time complexity and hyperparameter sensitivity, as it relies on $k$NN to determine the probability that each sample is noise. Second, AdaBoost directs base classifiers to focus on hard-to-classify samples by adjusting sample weights, whereas standard classifiers such as multilayer perceptron (MLP) inherently lack native support for sample weight-based mechanisms, thereby limiting the broader applicability of AdaBoost.

\begin{figure*}[h]
\centering
       \subfigure{
                \begin{minipage}[t]{0.45\linewidth}
                \centering
                \includegraphics[height=4.3cm, width=5.5cm]{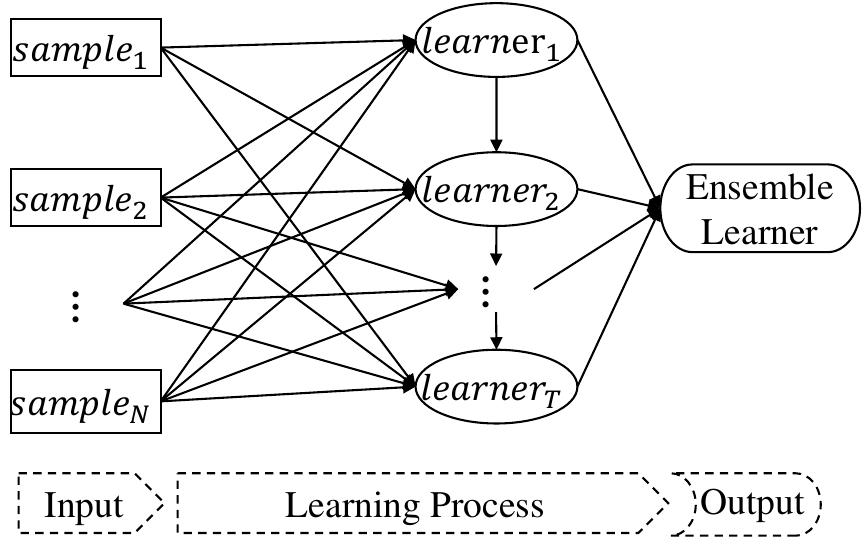}
                \\
                \small (a) Boosting with fine-grained input.
                \end{minipage}%
                        }%
        \subfigure{
                \begin{minipage}[t]{0.45\linewidth}
                \centering
                \includegraphics[height=4.3cm, width=5.5cm]{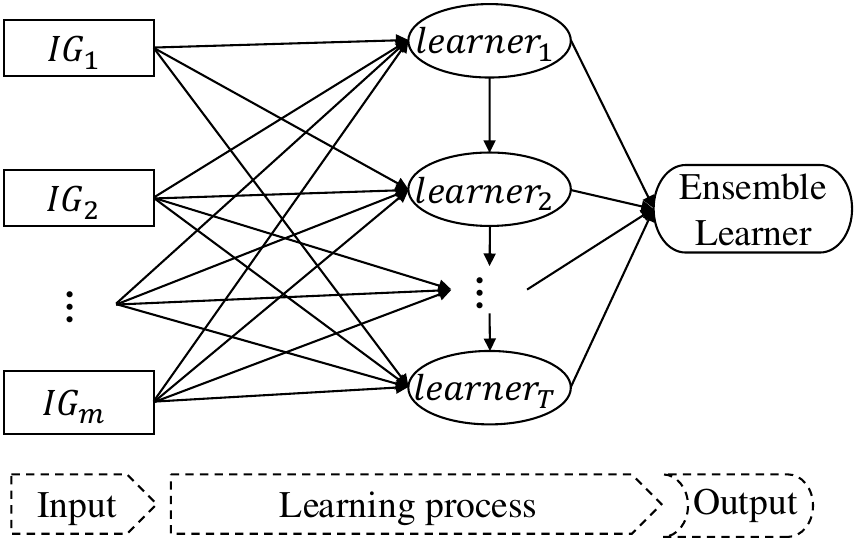}
                \\
                \small (b) Boosting with coarse-grained input.
                \end{minipage}%
                        }%
\caption{Comparison between traditional boosting algorithm and multi-granularity boosting algorithm.}
\label{fig1}
\end{figure*}

Third, to the best of our knowledge, as shown in Figure \ref{fig1}(a), existing AdaBoost algorithms use samples as units of computation (fine-grained inputs), which is not efficient. Granular computing (GrC) \cite{33}, also known as multi-granular computing, is a new paradigm that mimics the cognitive methods and information-processing characteristics of human beings. From the perspective of data analysis and processing, GrC can greatly improve computational efficiency by granulating complex data into information granules (IGs) and using IGs instead of samples as the basic unit of computation \cite{34}.
In recent years, inspired by the concept of GrC, many studies have advocated using IGs instead of samples as the input for machine learning algorithms, thereby improving their efficiency, robustness, and interpretability. Granular-ball computing (GBC) \cite{1} is one of its representative branches. For example, Figure \ref{fig1}(b) illustrates the boosting algorithm using coarse-grained input.

Inspired by GrC, this paper proposes a novel AdaBoost model framework called granular adaptive boosting (GAdaBoost). GAdaBoost is a two-stage learning framework consisting of a data granulation stage and an adaptive boosting stage. To demonstrate the feasibility of GAdaBoost, an extension of the SAMME algorithm within the framework, called GAdaBoost.SA, is presented. GAdaBoost.SA incorporates a GB generation (GBG) method and a GB-based SAMME algorithm. The main contributions of this work are mentioned below.

\begin{enumerate}[label= (\arabic*) , left=2em, itemsep=1pt] 
\item GAdaBoost is introduced. To the best of our knowledge, it is the first AdaBoost variant that implicitly updates the weights of IGs instead of individual samples, enabling it to focus on hard-to-classify objects, be base learner-agnostic, and achieve improved efficiency and robustness.
\item A GBG method tailored for ensemble learning is proposed. It efficiently compresses data while preserving diversity and remains robust to label noise, thereby ensuring the overall efficiency and robustness of GAdaBoost.
\item A SAMME extension within the GAdaBoost framework, namely the GAdaBoost.SA algorithm, is proposed. It retains the advantages of SAMME while leveraging the GB structure to enhance robustness and computational efficiency.
\end{enumerate}

The remainder of this paper is organized as follows. Section \ref{relwork} introduces related work, including the SAMME algorithm, its extension algorithm, and some basic knowledge about data granulation. Section \ref{GAdaBoost} introduces the proposed GAdaBoost framework. Section \ref{Proposed} presents the proposed GAdaBoost.SA algorithm. Section \ref{experiments} describes the numerical experiments conducted to validate the proposed model. Section \ref{Conclusion} concludes this work.

\section{Related Work}
\label{relwork}
This section first reviews the most popular multiclass AdaBoost algorithm, SAMME, and its extension for handling label noise, namely, Rob\_SAMME. Then, the principle of justifiable granularity (POJG) \cite{38,39} and the representative method of data granulation, namely, the GBG method, are introduced.
Table \ref{table1} summarizes the symbols commonly used in this paper.

\begin{table}[htbp]
    \caption{Notations and Descriptions.}
    \centering
    \label{table1}
    \footnotesize
    \renewcommand\tabcolsep{1.5pt}
    \begin{tabular}{cc}
        \hline
        \toprule[1pt]
        {Notation} & {Description} \\
        \hline
         $N$ & Number of samples \\
         $q$ & Number of features \\
         $K$ & Number of classes \\
         $D\in \mathbb R^{N \times q}$ & Training dataset \\
         $\chi\subseteq \mathbb R^q$ & Feature space \\
         $\mathcal{Y}=\{1,2,...,K\}$ & Label space \\
         $(\bm{x},y) \in D, \bm{x} \in \chi, y \in \mathcal{Y}$ & A sample \\
         $O$ & Sample set covered by an IG \\
         $p \in [0,1]$ & Purity of the label of the samples in $O$ \\
         $A$ & Ordered tuple of characteristics \\
         $\bm{c}$ & Center of a GB \\
         $\Omega$ & Distance set from the samples in the GB to $\bm{c}$ \\
         $g=(O,A)$,$gb=(O,(\bm{c},\Omega))$ & An IG and a GB \\
         $\mathbb G=\{g_i\}_{i=1}^{m^\prime}$,$G=\{gb_i\}_{i=1}^m$ & An IG set and a GB set \\
         $\gamma$ & Parameters of the base classifier \\
         $h(\bm{x};\gamma)$ & Base classifier \\
         $\beta$ & Weight of a base classifier \\
         $T$  & Number of base classifiers \\
        \hline
        \toprule[1pt]
    \end{tabular}
\end{table}

\subsection{SAMME and Robust Multiclass AdaBoost}
\label{adaboost}
SAMME \cite{17} (stagewise additive modeling using a multiclass exponential loss function) is a multi-class extension of AdaBoost. It retains AdaBoost’s core idea of improving classification performance by weighting samples and combining multiple base classifiers, forming what is known as an additive model (Eq.\ref{eq10}).
\begin{equation}\label{eq10}
f(\bm{x}) = \sum_{t=1}^T \beta_t h_t(\bm{x};\gamma_t).
\end{equation}

Given a training dataset $D$, the base classifier $h(\bm{x};\gamma)$, and an iteration number $T$, in the $t$th iteration, SAMME corrects the errors of the previous classifier $h_{t-1}(\bm{x};\gamma_{t-1})$ by training $h_t(\bm{x};\gamma_t)$ on $D$ with the updated sample weights, where weight of the misclassified samples are assigned higher weights than the correctly classified ones.
The difference from AdaBoost is that SAMME adopts a multiclass exponential loss function to handle both binary and multiclass classification scenarios (Eq.\ref{eq1}).
\begin{equation}\label{eq1}
L(y,f(\bm{x})) = \sum_{(\bm{x},y)\in D} \exp\left(-\frac{1}{K}\sum_{k=1}^K \mathbb I(y=k)f^k(\bm{x})\right),
\end{equation}
where $f^k(\bm{x})$ represents the cumulative voting score for class $k$, $\mathbb I (\cdot)$ represents the indicator function, and $k\in \mathcal{Y}$ represents the class.

The goal of SAMME is to minimize the loss function (Eq.\ref{eq2}).
\begin{equation}\label{eq2}
\arg\min_{\beta_t, \gamma_t}\sum_{i=1}^N L(y_i,\sum_{t=1}^T \beta_t h_t(\bm{x}_i; \gamma_t)).
\end{equation}
SAMME employs the same weak learnability criterion as AdaBoost, requiring the base classifier to perform slightly better than random guessing. As a result, SAMME adjusts the error rate tolerance for the base classifier from $\epsilon < 0.5$ to $\epsilon < 1-1/K$ to accommodate multi-class classification scenarios.
Therefore, the weight $\beta_t$ of the $h_t(\bm{x};\gamma_t)$ can be derived by solving Eq.\ref{eq2}, as shown in Eq.\ref{eq3}.
\begin{equation}\label{eq3}
\beta_t = \ln(\frac{1-\epsilon_t}{\epsilon_t}) + \ln(K-1).
\end{equation}

The main steps of SAMME are summarized as follows. The weights of the samples in $D$ are initialized as $w_i^1=1/N,i=1,2,...,N$. Step 1: Train the $t$th base classifier $h_t(\bm{x};\gamma_t)$ on $D$ using the weights $w_i^t$. Step 2: Compute the error rate $\epsilon_t$ of $h_t(\bm{x};\gamma_t)$, $\epsilon_t=\sum_{i=1}^N w_i^t\mathbb{I}\left(y_i\neq h_t(\bm{x}_i;\gamma_t)\right)/\sum_{i=1}^N w_i^t$. Step 3: Compute the weights $\beta_t= \ln(\frac{1-\epsilon_t}{\epsilon_t}) + \ln(K-1)$ of $h_t(\bm{x};\gamma_t)$. Step 4: Update the sample weights, $w_i^{t+1}\leftarrow w_i^{t}\cdot\exp\left(\beta_t\cdot\mathbb{I}\left(y_i\neq h_t(\bm{x}_i;\gamma_t)\right)\right)$, and normalize the weights. Step 5: Repeat steps 1-4 until $T$ base classifiers are trained. Finally, the strong classifier is shown as Eq.\ref{eq4}.
\begin{equation}\label{eq4}
F(\bm{x})=\arg\max_k \sum_{t=1}^T \beta_t \times \mathbb I (h_t(\bm{x};\gamma_t)=k).
\end{equation}

Rob\_SAMME \cite{32} enhances the robustness of SAMME by introducing a noise detection function (NDF) and a new sample weight update scheme. The main idea of NDF is as follows. In the $t$th iteration, the base classifier $h_t(\bm{x};\gamma_t)$ is trained, and then NDF is employed to evaluate whether $\forall (\bm{x}_i,y_i) \in D$ is a noise sample. Specifically, the neighborhood error rate $\mu_t(\bm{x}_{i},y_{i})=\frac{1}{k}\sum_{j=1}^{k}\mathbb I(h_t(\bm{x}_j;\gamma_t)\neq y_{j})$, $(\bm{x}_j,y_j)\in neighbors(\bm{x}_i,y_i)$, of each sample $(\bm{x}_i,y_i)$ is calculated. If $\mu_t(\bm{x}_i,y_i)$ is higher than the average neighborhood error rate of all samples $\overline{\mu_t}=\frac{1}{N}\sum_{i=1}^{N}\mu_t (\bm{x}_{i},y_{i})$, $(\bm{x}_i,y_i)$ is considered to be a noise sample; otherwise, it is considered a non-noise sample.
In addition, Rob\_SAMME determines whether $h_t(\bm{x};\gamma_t)$ should be discarded based on the error rate $\epsilon_t$ of $h_t(\bm{x};\gamma_t)$. If $\epsilon_t>(K-1)/K$ or $\epsilon_t=0$, the sample weights are reset, that is, $w_{i}^{t}= 1/N$, and $h_t(\bm{x};\gamma_t)$ is retained.
The core idea of the new sample weight update scheme is to update according to the performance of $h_t(\bm{x};\gamma_t)$ on non-noise samples. Specifically, in the $t$th iteration, the weight of a misclassified non-noise sample is increased by $w_{i}^{t+ 1}\leftarrow w_{i}^{t}\times \exp (\beta _{t})$, $\beta_{t}=\frac{(K-1)^{2}}{K}\left(ln\left(\frac{1-\epsilon_t}{\epsilon_t}\right)+ln\left(K-1\right)\right)$, whereas the weight of a correctly classified noise sample is set to zero.
However, each iteration of Rob\_SAMME requires computing the $k$ nearest neighbors of each sample to identify noise, making it time-consuming and sensitive to the parameter $k$.

\subsection{Granulation}
\label{Granulation}
Data granulation decomposes complex data into IGs according to a given granulation strategy, essentially serving as an effective data compression method. Various granulation strategies can be adopted depending on different scenario requirements.
Pedrycz \cite{41,42} proposed POJG to unify the granulation standards, namely, \textit{Coverage} and \textit{Specificity}. \cite{43} pointed out that the two criteria are mutually exclusive. Therefore, a multiobjective optimization model can be designed to determine the optimal granularity for constructing IGs. Since the introduction of POJG, numerous studies have been conducted based on it, including super interval granular rough sets \cite{44}, heterogeneous integration at an appropriate granularity \cite{35}, and interval-valued time series prediction model at an appropriate granularity \cite{45}, etc.

\begin{itemize}[left=2em, itemsep=1pt] 
\item \textbf{Coverage:} The IG should contain as much data as possible to ensure that its existence is well justified. For example, if an IG is a sample set, then IG $g_1=\{(\bm{x}_1,y_1),(\bm{x}_2,y_2),(\bm{x}_3,y_3)\}$ has more legitimate experimental evidence than IG $g_2=\{(\bm{x}_1,y_1),(\bm{x}_2,y_2)\}$.
\item \textbf{Specificity:} The generated IG should have well-defined semantics. This implies that the more compact the IG, the better. For example, if an IG is a sample set, then the IG $g_1=\{(\bm{x}_1,y_1),(\bm{x}_2,y_2)\}$ has clearer semantics than the IG $g_2=\{(\bm{x}_1,y_1),(\bm{x}_2,y_2),(\bm{x}_3,y_3)\}$, where $y_3\neq y_1,y_2$.
\end{itemize}

GBC \cite{1,2} has emerged as a vital branch of GrC in recent years. It uses GBs of different sizes to adaptively represent and cover the sample space and learn based on these GBs. The GBG method of GBC is an efficient and robust granulation method that aligns with POJG. The basic steps of GBG are as follows:
First, the entire $D$ is initialized to a GB $gb_0$. Next, the $gb_0$ is divided into $k$ or more smaller GBs using methods such as $k$-means \cite{1}, $k$-division \cite{2}, and hard-attention division \cite{47}. The center $\bm{c}_i$ and radius $r_i$ of $\forall gb_i \in G$ are defined in Definition \ref{mydef1}.

\begin{mydef}\label{mydef1}\cite{1} Given a dataset $D$, let $G$ be a set of GBs generated on $D$. For $\forall gb_i\in G$, it is generated on $O_i (O_i \subseteq D)$, and the center $\bm{c_i}$ and radius $r_i$ of $gb_i$ are respectively defined as Eq.\ref{eq5} and \ref{eq6}.
\begin{equation}\label{eq5}
\bm{c}_i=\frac{1}{|O_i|}\sum_{(\bm{x},y)\in O_i}\bm{x},
\end{equation}
\begin{equation}\label{eq6}
r_i=\frac{1}{|O_i|}\sum_{(\bm{x},y) \in O_i} \bigtriangleup(\bm{x},\bm{c_i}),
\end{equation}
where $|\bullet|$ represents the cardinality of set $\bullet$ and $\bigtriangleup(\cdot)$ denotes the distance function. Without losing generality, distances in this paper refer to Euclidean distances.
\end{mydef}

The quality of the $gb_i$ is measured by its purity $p_i$, as defined in Definition \ref{mydef2}. The closer the purity of a GB to $1.0$, the more closely its distribution aligns with that of the original dataset.

\begin{mydef}\label{mydef2}\cite{1} Given a dataset $D$, let $G$ be a set of GBs generated on $D$. For $\forall gb_i \in G$, it is generated on $O_i (O_i \subseteq D)$, and its label $l_i$ and purity $p_i$ are defined respectively as Eq.\ref{eq7} and \ref{eq8},
\begin{equation}\label{eq7}
l_i={ \underset{k \in \mathcal{Y}}{{\arg\max} \,} |\{(\bm{x},y)\in O_i|y=k\}|},
\end{equation}
\begin{equation}\label{eq8}
p_i=\frac{|\{(\bm{x},y)\in O_i|y=l_i\}|}{|O_i|}.
\end{equation}
\end{mydef}

Each $gb_i$ is iteratively split until its purity reaches a given threshold. Finally, a GB set $G$ is obtained. $G$ has been verified to approximately describe the original dataset $D$. However, the primary aim of the existing GBG method is to serve as a general data granulation method. In certain scenarios, such as ensemble learning, this lack of specificity can lead to unsatisfactory performance and a waste of resources, such as computational inefficiency.

\section{Granular AdaBoost Model}
\label{GAdaBoost}
The essence of AdaBoost lies in dynamically adjusting sample weights to focus on misclassified samples, thereby enhancing overall learning performance. Inspired by this mechanism, GAdaBoost shifts the focus from individual samples to coarse-grained IGs. By implicitly and dynamically updating the weights of IGs containing misclassified samples, GAdaBoost effectively improves learning performance while leveraging the advantages of GrC.

\begin{figure*}[ht]
\centering
\includegraphics[width=15cm]{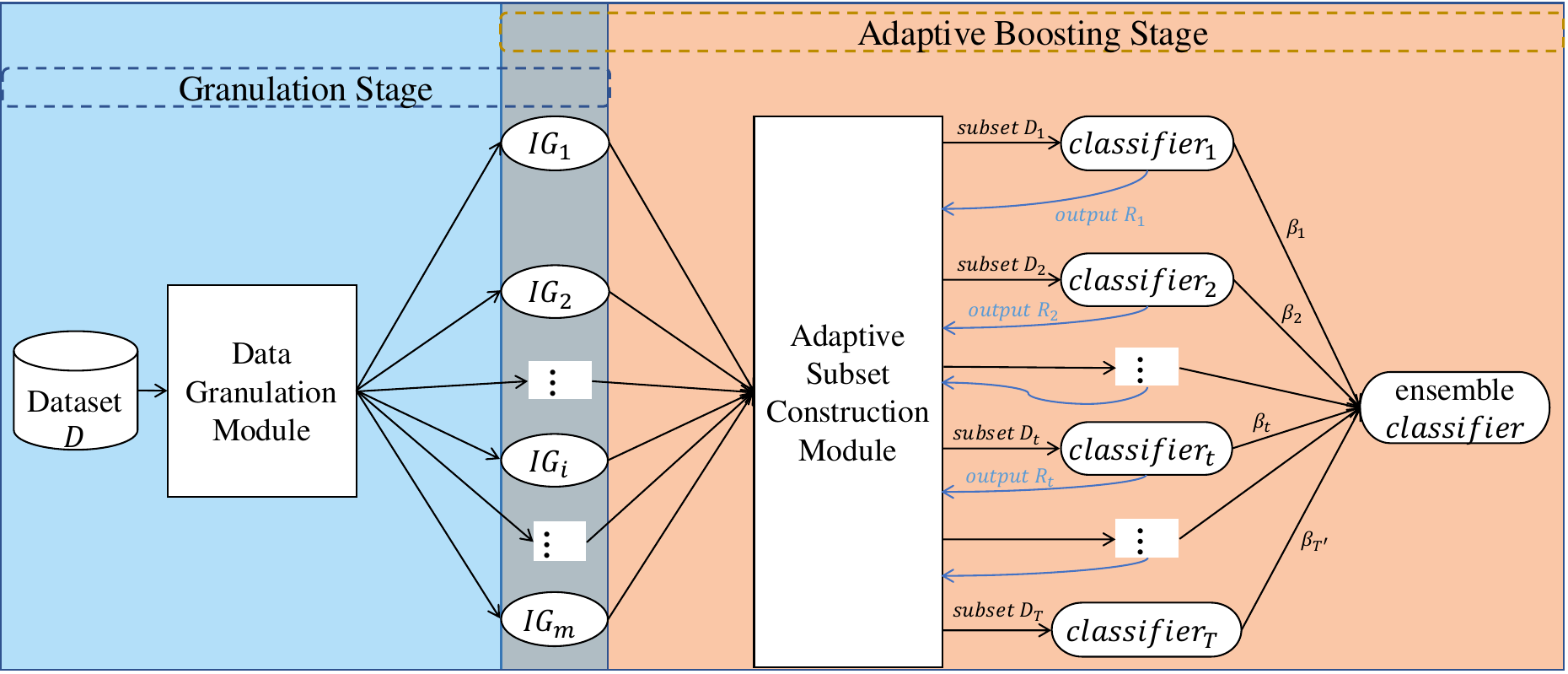}
\caption{The Framework of GAdaBoost}
\label{fig2}
\end{figure*}

This section introduces GAdaBoost in detail, with its framework shown in Figure \ref{fig2}.
As illustrated in Figure \ref{fig2}, GAdaBoost is a two-stage learning model consisting of a granulation stage and an adaptive boosting stage based on IGs. First, justifiable granulation methods transform finest-grained data into coarse-grained IGs. Next, an adaptive subset construction module constructs a training subset from IGs for the next base classifier, adaptively guided by the previous base classifier’s outputs, thereby implicitly adjusting the IG weights. Finally, by combining a sequence of base classifiers, GAdaBoost produces a strong classifier. These steps enable GAdaBoost to efficiently leverage IGs for improved performance. The main contributions of GAdaBoost are summarized as follows:

\begin{enumerate}[label= (\arabic*) , left=2em, itemsep=1pt] 
\item GAdaBoost shifts the focus from individual samples to IGs by dynamically and implicitly updating their weights, placing greater emphasis on hard-to-classify IGs. This coarse-grained perspective enhances the computational efficiency and scalability of AdaBoost in complex data scenarios.
\item Unlike traditional AdaBoost, GAdaBoost iteratively trains base classifiers on dynamically constructed subsets of the training dataset, significantly reducing training time and memory consumption while maintaining model performance.
\item The granulation stage in GAdaBoost leverages the coarse-grained representation of data to mitigate the impact of label noise, thereby enhancing the robustness of AdaBoost in label noise environments.
\end{enumerate}

\subsection{POJG for Ensemble Learning}
\label{POJG}
As discussed in Section \ref{Granulation}, the essence of POJG is to construct IGs with intrinsic value based on existing data. Building on the POJG, this section introduces the principle of justifiable granularity for ensemble learning (POJG-Ens).

Ensemble learning focuses on generating ``good and diverse" base learners to improve generalization performance. Most boosting algorithms achieve diversity by adjusting the data distribution. Traditional ensemble learning typically uses individual samples as input, which inherently provides maximal diversity at the input level. When IGs replace individual samples as input, ensuring that the constructed IGs also maintain sufficient diversity is essential. To address this, POJG is extended by incorporating a diversity criterion, leading to the POJG-Ens, stated as follows.

\begin{itemize}[itemsep=1pt]
\item \textbf{Coverage:} As stated in the POJG.
\item \textbf{Specificity:} As stated in the POJG.
\item \textbf{Diversity:} The differences between IGs should be as large as possible to ensure that the base learners explore the data from diverse perspectives.
\end{itemize}

\begin{example}\label{example1}
Given $D=\{(\bm{x}_1,y_1), (\bm{x}_2,y_2), (\bm{x}_3,y_3), (\bm{x}_4,y_4), (\bm{x}_5,y_5), (\bm{x}_6,y_6)\}$, if there are two granulation results, namely $\mathbb G_1=\{\{(\bm{x}_1,y_1), (\bm{x}_2,y_2)\}, \{ (\bm{x}_3,y_3), (\bm{x}_4,y_4)\},\{(\bm{x}_5,y_5), (\bm{x}_6,y_6)\}\}$ and $\mathbb G_2=\{\{(\bm{x}_1,y_1), (\bm{x}_2,y_2), (\bm{x}_3,y_3)\},\{(\bm{x}_4,y_4), (\bm{x}_5,y_5), (\bm{x}_6,y_6)\}\}$, where each IG is a sample set. From the perspective of diversity, $\mathbb G_1$ provides a more diverse granulation than $\mathbb G_2$, as its IGs are more fine-grained, leading to greater differentiation among them.
\end{example}

Evidently, \textit{Coverage}, \textit{Specificity}, and \textit{Diversity} are mutually constrained. When designing a granulation method, it is crucial to consider how to balance them. Although \textit{Specificity} can be quantified in various ways, we adopt label consistency as its measure. Based on this, the mathematical model is formulated as follows:

Given a training dataset $D$, assume that by using the justifiable granulation method, $D$ can be transformed into an IG set $\mathbb G$. Let us denote IG as $g=(O,p)$, where the ordered tuple of characteristics $A$ has only one element, namely, $p$. The loss function can be expressed as $L_{gran}(\mathbb G) = \alpha_1\times l_c(\mathbb G) + \alpha_2 \times l_s(\mathbb G) + \alpha_3 \times l_d(\mathbb G)$. The objective function of the granulation method can be expressed as Eq.\ref{eq9}.
\begin{equation}
\label{eq9}
\begin{aligned}
\mathbb G^* = \arg\min_{\mathbb G} L_{gran}(\mathbb G) \\
\text{s.t.}
\left\{\begin{matrix}
 {\textstyle \bigcup_{i=1}^{m}}  O_i \subseteq D, \\
 O_i \bigcap O_j = \emptyset, \forall i \neq j,\\
 \alpha_1 +\alpha_2 +\alpha_3 =1,\\
\alpha_1,\alpha_2,\alpha_3 \ge 0,
\end{matrix}\right.
\end{aligned}
\end{equation}
where $l_c(\mathbb G) = \sum_{i=1}^{m^\prime} \varphi (O_i)$, with $\varphi(\cdot)$ being a monotonically decreasing function; $l_s(\mathbb G) = \sum_{i=1}^{m^\prime} \psi(p_i)$, where $\psi(\cdot)$ is also a monotonically decreasing function; and $l_d(\mathbb G) = \sum_{i,j=1}^{m^\prime} \eta (s(g_i, g_j))$, $i\ne j$, where $s(g_i, g_j)$ represents the similarity measurement between $g_i$ and $g_j$, and $\eta (\cdot)$ is a monotonically increasing function. Furthermore, $\alpha_1,\alpha_2,\alpha_3$ represent the coefficients of each criterion.

\subsection{GAdaBoost}
\label{subsec1}

As previously stated, GAdaBoost dynamically adjusts the attention allocated to these IGs during each iteration, ensuring that subsequent base classifiers focus on learning the challenging IGs that contain misclassified samples. This refinement in classification performance on these IGs enhances the overall accuracy with each iteration. This approach aligns with the additive model framework in Eq.\ref{eq10}, thereby leveraging its strengths and extending its capabilities.

The IG set $\mathbb G$, obtained using the granulation method designed according to POJG-Ens proposed in Section \ref{POJG}, is utilized as the input. GAdaBoost constructs “good and different” base classifiers by generating a diverse subset of the training dataset on $\mathbb G$. This strategy ensures that each base classifier learns from a unique perspective. As illustrated in Figure \ref{fig2}, the primary steps of GAdaBoost can be outlined as follows:

\begin{enumerate}[label=Step \arabic*. , left=2em,itemsep=0.5pt] 
\item Convert $D$ to $\mathbb G$ using a reasonable granulation method.
\item Construct the initial training subset $D_1(D_1\subset D)$ on $\mathbb G$.
\item Train the $t$th base classifier $h_t(\bm{x};\gamma_t)$ on $D_t$ to obtain its classification result $R_t$ and weight $\beta_t$.
\item Construct the training subset $D_{t+1}(D_t\subset D_{t+1}, D_{t+1}\subseteq D)$ according to the guidance of $R_t$.
\item Repeat Steps 3 and 4 until the given $T$ base classifiers are obtained or the training subset $D_t$ meets the stop condition.
\end{enumerate}

Given the training dataset $D$ and the loss function $L(y, f(\bm{x}))$, the classical SAMME algorithm optimizes the additive model in Eq.\ref{eq10} using empirical risk minimization, as formulated in Eq.\ref{eq2}. Building on this framework, GAdaBoost introduces an adaptive refinement strategy, leading to a modified optimization problem (Eq.\ref{eq11}).
\begin{equation}\label{eq11}
\arg\min_{\beta_t, \gamma_t}\sum_{i=1}^{N^{\prime}} L(y_i,\sum_{t=1}^{T^\prime} \beta_t h_t(\bm{x}_i; \gamma_t)),
\end{equation}
where $N^\prime \le N$, $T^\prime \le T$. Eq.\ref{eq11} reflects the selective learning mechanism in GAdaBoost, where both the number of training samples and the number of base classifiers are adaptively adjusted. The corresponding total loss function is then defined as Eq.\ref{eq12}.

\begin{equation}\label{eq12}
\begin{aligned}
L_{pred}(y,\sum_{t=1}^{T^\prime} \beta_t h_t(\bm{x}; \gamma_t)) + \delta \times L_{gran}(\mathbb G),
\end{aligned}
\end{equation}
where $\delta$ is the balance coefficient.
In summary, the objective function is as Eq.\ref{eq13}.
\begin{equation}\label{eq13}
\begin{aligned}
\arg\min_{\mathbb G,\beta,\gamma} (L_{pred}(y,\sum_{t=1}^{T^\prime} \beta_t h_t(\bm{x}; \gamma_t)) + \delta \times L_{gran}(\mathbb G)).
\end{aligned}
\end{equation}

\section{Proposed Approach}
\label{Proposed}
\begin{figure*}[ht]
\centering
\includegraphics[width=16cm]{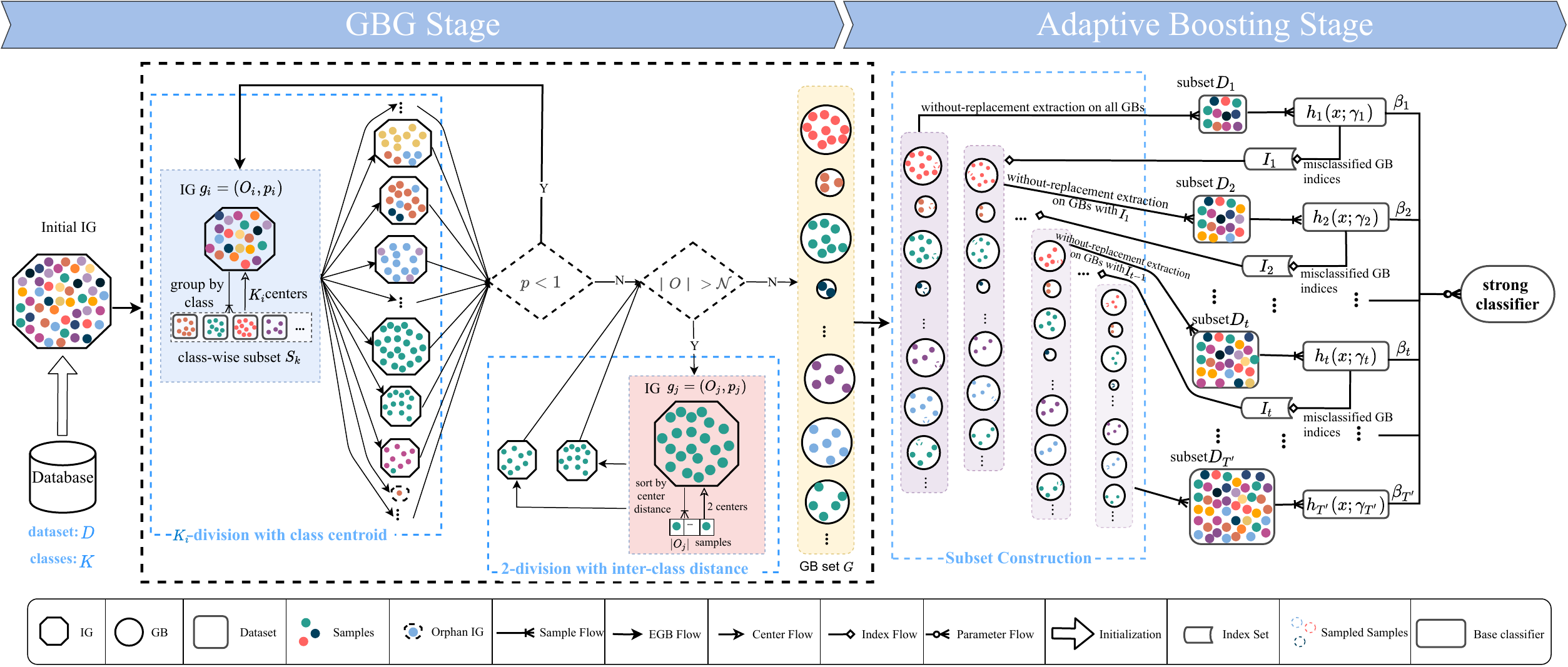}
\caption{The Framework of GAdaBoost.SA.}
\label{fig3}
\end{figure*}

In this section, GAdaBoost.SA, an extension of SAMME designed for label noise scenarios, is proposed to demonstrate the feasibility of GAdaBoost. As illustrated in Figure \ref{fig3}, GAdaBoost.SA adopts a two-stage learning approach comprising a GBG stage and an adaptive boosting stage. In the first stage, a GBG method tailored for ensemble learning (GBG-Ens) is developed based on POJG-Ens to transform $D$ to $G$. In the second stage, $G$ serves as the input, and the training subset $G_t$ is iteratively constructed on the GBs (index set $I_t$) containing misclassified samples from a previous base classifier. A series of base classifiers $h_t(\bm{x}; \gamma_t)$ are trained and weighted based on their ability $\beta_t$ to obtain a strong classifier. Notably, an early stopping mechanism is triggered when the training subset converges.

The motivation behind the proposed GAdaBoost.SA algorithm is as follows:
\begin{enumerate}[label= (\arabic*) , left=2em, itemsep=1pt]
\item Existing granulation methods, such as the GBG method, often overlook the diversity criterion essential for constructing effective ensemble models. This limitation may result in suboptimal support for training multiple ``good and different" base classifiers.
\item SAMME trains base classifiers on the entire training dataset and lacks a mechanism to handle redundant or label noise, leading to low computational efficiency and a tendency to overfit on noisy datasets.
\item Existing SAMME extension for handling label noise relies on $k$NN in each iteration, which brings challenges such as sensitivity to hyperparameters and high computational cost.
\end{enumerate}

\subsection{GBG-Ens}
\label{CEGB}
Solving the multi-objective optimization problem formed by considering POJG-Ens is computationally expensive because of the complexity of simultaneously optimizing coverage, specificity, and diversity. Therefore, inspired by GBC, we approximate the solution to this problem using heuristic methods to balance efficiency and effectiveness.

The symmetric structure of GBs facilitates identifying edge samples in local clusters formed by GB samples. To introduce the GBG-Ens method, the extended GB is first defined.

\begin{mydef}\label{mydef3}
Given a dataset $D$, let $G=\bigcup gb$ be a set of GBs generated on $D$. For $\forall gb\in G$, $gb=(O,(\bm{c},\Omega))$, it is generated on $O (O \subseteq D)$, and the radial distance set $\Omega$ is defined as Eq.\ref{eq14}.
\begin{equation}\label{eq14}
\Omega= \{d_{i}\mid d_{i}=\bigtriangleup(\bm{x_i},\bm{c}),(\bm{x_i},y_i)\in O\}.
\end{equation}
In other words, GB is a special case of IG when the ordered tuple of characteristics $A$ is $(\bm{c}, \Omega)$.
\end{mydef}

According to Eq.\ref{eq6}, there is $max(\Omega)=r$. In other words, the samples farther from $\bm{c}$, that is, $(\bm{x_i},y_i)\in O$ with larger $d_{i}$, are more likely to be distributed near the surface of $gb$.

During the granulation, because the GB structure is unnecessary, the IG $g = (O,p)$ is used to represent coarse-grained objects to avoid redundant computations. Figure \ref{fig3} (left) outlines the proposed GBG-Ens method. First, the entire training dataset $D$ is initialized as a single IG. Then, a $k$-division method is iteratively applied to IGs that do not meet the POJG-Ens conditions, progressively refining them until all IGs satisfy the criteria. Finally, each IG is assembled into the GB, resulting in the GB set.

Algorithm \ref{alg1} provides a detailed description of the proposed GBG-Ens method. Given a training dataset $D$, initialize $D$ to an IG $g_1^1=(O_1^1,p_1^1)$, and set the IG set $\mathbb G =\{g_1^1\}$. Because $p\in[0,1]$, a larger $p$ indicates a higher score for $g$ under the \textit{Specificity} criterion. Therefore, the \textit{Specificity} criterion can be formalized as the requirement that $p=1$. Because $K\ge2$, there is $p_1^1<1$ for $g_1^1$. At this point, the score of the \textit{Coverage} criterion reaches the optimal level, whereas the \textit{Specificity} and \textit{Diversity} remain poor. According to Section \ref{POJG}, the objective function as in Eq.\ref{eq9} cannot reach the optimal level.
Therefore, $\forall g_i^j\in \mathbb G$ with $p_i^j < 1$ will be iteratively split. The splitting process can be briefly described as follows:

\begin{enumerate}[label=Step \arabic*. , left=2em] 
\item Group the samples in $O_i^j$ by label category to obtain ${\textstyle \bigcup_{k=1}^{K_i^j}}S_k=O ,K_i^j \le K$, and calculate the mean feature vector of $S_k$ as the new center $\bm{c}_{i+1}^k$ using Eq.\ref{eq5}.
\item Compute the distance from $\forall (x,y)\in O_i^j$ to every center $\bm{c}_{i+1}^k$ and assign $(x,y)$ to the nearest center to form the new IG $g_{i+1}^k$. Then, obtain the new IG set $\bigcup g_{i+1}^k$, where $\left | \bigcup g_{i+1}^k \right |\le K_i^j$.
\item Determine the rationality of the split result. If $\left | \bigcup g_{i+1}^k \right |=1$, the distribution of different classes in $O_i^j$ overlaps significantly, or $O_i^j$ may contain label noise. To reduce noise interference and simplify model training, non-majority class samples in $g_{i+1}^k$ are removed.
\item Merge the new IG set $\bigcup g_{i+1}^k$ into $\mathbb G$, $\mathbb G=\mathbb G+\bigcup g_{i+1}^k$.
\end{enumerate}

\begin{algorithm}[H]
\caption{GBG-Ens Method.}
\label{alg1}
\SetKwInOut{Input}{Input}\SetKwInOut{Output}{Output}
\Input {Dataset $D$.}
\Output {GB set $G$.}
Treat the entire $D$ as the initial IG $g_i^j=(O_i^j,p_i^j)$, $i=1,j=1$\;
Initialize $\mathcal{N}=\left \lfloor \frac{1}{K-1}\sqrt{N} \right \rfloor$, $G\leftarrow \emptyset$, $\mathbb G \leftarrow \{g_1^1\}$ \;
\Repeat{$\mathbb G==\emptyset$}{
    \For {each $g_i^j \in \mathbb G$}{
        \If {$|O_i^j| \le 1$}{
            $\mathbb G \leftarrow \mathbb G - \{g_i^j\}$ \;
            \textbf{\texttt{continue}}\;
        }
        Calculate purity $p_i^j$ using Eq.\ref{eq8} \;
        \If {$p_i^j \le 1$}{
            Group $O_i^j$ by class to obtain ${\textstyle \bigcup_{k=1}^{K_i^j}}S_k=O_i^j$\;
            Calculate the centroid $\bm{c}_{i+1}^k$ of class-wise subset $S_k$ using Eq.\ref{eq5}\;
            Split $g_i^j$ into $\bigcup g_{i+1}^k$ based $\bm{c}_{i+1}^k$ using $k$-division \;
            \If {$|\bigcup g_{i+1}^k|==1$}{
                Delete samples in $O_{i+1}^1$ that are inconsistent with majority class\;
                $\mathbb G \leftarrow \mathbb G - \{g_i^j\}$\;
                $\mathbb G \leftarrow \mathbb G + \bigcup g_{i+1}^k$\;
            }
        }
        \ElseIf {$|O_i^j| > \mathcal{N}$}{
            Take samples closest and farthest from the center $c_i^j$ of $O_i^j$ as the new center $\bm{c}_{i+1}^k,k=1,2$\;
            Split $g_i^j$ into $\bigcup_{k=1}^2 g_{i+1}^k$ based $\bm{c}_{i+1}^k$ using $2$-division\;
            $\mathbb G \leftarrow \mathbb G - \{g_i^j\}$\;
            $\mathbb G \leftarrow \mathbb G + \bigcup g_{i+1}^k$\;
        }
        \Else{
            Construct GB $gb$ on $O_i^j$ according to Definition \ref{mydef3}\;
            $G\leftarrow G+\{gb\}$\;
            $\mathbb G \leftarrow \mathbb G - \{g_i^j\}$\;
        }
    }
}
\textbf{return} $G$.
\end{algorithm}

Iteratively perform the above process until all IGs $g\in \mathbb G$ satisfy $p=1$. This results in an IG set $\mathbb G$ with the highest \textit{Specificity} score and relatively good \textit{Coverage} score. At this time, it is necessary to further consider \textit{Diversity} comprehensively to balance the trade-off between \textit{Coverage} and \textit{Diversity}.

In unsupervised learning, the number of clusters is often constrained by factors such as data dimensionality and distribution complexity, and does not grow linearly. Some works \cite{3,4,5} suggest a rule of thumb where the maximum number of clusters is set to $\sqrt{N}$, balancing the trade-off between overfitting due to excessive clustering and under-segmentation due to insufficient clusters. Inspired by this, a scaling factor $\mu$ is introduced to adapt to different scenarios, yielding a maximum cluster count of $\mu \times \sqrt{N}$. For supervised learning tasks, where category granularity is a consideration, the maximum sample capacity covered by an IG is set to $\mathcal{N} = \left \lfloor \frac{1}{K-1} \sqrt{N} \right \rfloor$.
Following the similarity-within and difference-between principle in unsupervised learning, for $\forall g_i^j \in \mathbb G$ with $p_i^j=1$, if $\left | O_i^j \right | \geq \mathcal{N}$, it is divided into two finer IGs. The main steps of the division process are outlined as follows:

\begin{enumerate}[label=Step \arabic*. , left=2em]
\item Based on the distance between $\forall (\bm{x},y) \in O_i^j$ and its center $\bm{c}_i^j$, take the closest and farthest samples as the new centers $\bm{c}_{i+1}^k,k=1,2$.
\item Assign each sample $(\bm{x},y)\in O_i^j$ to its nearest center $\bm{c}_{i+1}^k$, forming two finer IGs $\bigcup g_{i+1}^k$.
\item Merge the new IG set $\bigcup g_{i+1}^k$ into $\mathbb G$, $\mathbb G=\mathbb G+\bigcup g_{i+1}^k$.
\end{enumerate}

It is evident that, by following the aforementioned steps, the IG set $\mathbb G$ can be obtained with an optimal \textit{Specificity} score and balanced \textit{Coverage} and \textit{Diversity} scores. Notably, if a sample forms an IG in isolation, it is likely to be either an outlier or label noise. Therefore, any IG that contains only one sample will be removed.

In the GAdaBoost framework, the data subset used to train the base classifier should have information comparable to the original dataset, such as class boundary information. To achieve this, according to Definition \ref{mydef3}, the IG set $\mathbb G$ is transformed into a GB set $G$. The GB set will be utilized in Section \ref{GAdaBoost.SA} to construct training subsets that effectively capture class boundary samples.

\subsection{GAdaBoost.SA}
\label{GAdaBoost.SA}

Algorithm \ref{alg2} provides a detailed description of the proposed GAdaBoost.SA.
First, it considers how to train a ``good" base classifier. Based on POJG-Ens, the distribution of the constructed GB set approximates that of the original dataset. Leveraging the geometric properties of the ball, samples farther from the center are more likely to be distributed near the surface. If a ball overlaps the class boundary, samples near its surface will likely be boundary samples.
Moreover, in the GAdaBoost framework, relying solely on the boundary samples of each ball to train the first base classifier may deliver a strong initial performance but lack sustainability across subsequent iterations. Therefore, the initial training subset should balance sufficient sample size and representation. The initial sampling strategy is designed as follows. Empirical evidence in Boosting suggests that using approximately half of the dataset often yields good performance. Considering dataset characteristics such as scale and sparsity, $min(\left \lfloor 0.5*\mathcal{N} \right \rfloor,q)$ samples are selected from each GB to construct the initial training subset $D_1$. Fewer samples are often sufficient for low-dimensional datasets with simpler distributions to train a ``good" base classifier, reducing redundant computations while maintaining performance.

Second, consider how to train a ``different" base classifier. Based on the core idea of AdaBoost, the $t$th base classifier should pay more attention to the misclassified samples of the $(t-1)$th base classifier, $t>1$. In the GAdaBoost framework, because of the compactness of samples within GB, the misclassification of any sample $(\bm{x},y)$ suggests that other samples within the same GB may also be misclassified.
Thus, the subset $D_t$ is constructed by incrementally adding unsampled samples from GBs containing misclassified samples of the $(t-1)$th base classifier to enhance diversity among base classifiers. The steps for the incremental sampling strategy are as follows:
\begin{enumerate}[label= (\arabic*) ]
\item
Obtain the index set $I_{t-1}$ of the GBs containing misclassified samples from the $(t-1)$th base classifier.
\item Sample from misclassified GBs as follows:
\begin{enumerate}[label= \arabic*) ,left=1em]
\item If $|I_{t-1}| \neq 0$, from each GB in $I_{t-1}$, select the farthest unsampled sample from the center and add it to the training subset $D_{t-1}$, forming $D_t$.
If $|D_{t-1}|==|D_t|$, meaning all samples in the GBs corresponding to $I_{t-1}$ have been collected, stop the iteration as the training subset has converged.
\item If $|I_{t-1}| = 0$, conclude that the error rate of the $(t-1)$th base classifier is $0$, and stop the iteration.
\end{enumerate}
\end{enumerate}

Because GAdaBoost.SA adjusts the distribution of the data by implicitly modifying the weight of GBs instead of samples, the error rate $\epsilon_t$ of $h_{t}(\bm{x};\gamma_{t})$ is defined as Eq.\ref{eq15}.

\begin{equation}\label{eq15}
\epsilon_t = \frac{1}{|D_{t}|} \sum_{(\bm{x},y) \in D_t} \mathbb I(h_{t}(\bm{x};\gamma_{t}) \neq y).
\end{equation}

The weight $\beta_t$ of the $h_t(\bm{x},\gamma_t)$ is consistent with Eq.\ref{eq3}.
Theoretically, it will be analyzed in Section \ref{Justification} that the weight defined in Eq.\ref{eq15} can also make the multi-class exponential loss function in Eq.\ref{eq1} converge.

Finally, the strong classifier $F(x)$ can be defined as follows.

\begin{equation}\label{eq16}
F(\bm{x})=\arg\max_k \sum_{t=1}^{T^\prime} \beta_t \times \mathbb I (h_t(\bm{x};\gamma_t)=k).
\end{equation}
Because the subset $D_t$ gradually converges, $\epsilon_t=0$ may occur. Therefore, $T^{\prime}\le T$; in other words, GAdaBoost.SA inherits the early stopping mechanism.

\begin{algorithm}[htbp]
\caption{GAdaBoost.SA.}
\label{alg2}
\SetKwInOut{Input}{Input}\SetKwInOut{Output}{Output}
\Input {Dataset $D$, base classifier $h(\bm{x},\gamma)$, number of
iterations $T$.}
\Output {Strong classifier $F(\bm{x})$.}
Run Algorithm 1 to granulate the dataset $D$ into a GB set $G$\;
Take out $min(\left \lfloor 0.5*\mathcal{N} \right \rfloor,q)$ samples farthest from the center of each $gb\in G$ to form the initial training subset $D_1$\;
\For {$t=1,2,...,T$}{
    Train the base classifier $h_t(\bm{x},\gamma_t)$ on the training subset $D_t$\;
    Calculate the error rate $\epsilon_t$ of $h_t(\bm{x},\gamma_t)$ on $D_t$ using Eq.\ref{eq15}\;
    Calculate the weight $\beta_t$ of $h_t(\bm{x},\gamma_t)$ using Eq.\ref{eq3}\;
    Obtain the index set $I_t$ of the GB where the misclassified samples of $h_t(\bm{x},\gamma_t)$ are located\;
    \If {$|I_i| \neq 0$}{
        In each corresponding GB of $I_t$, take an unsampled sample farthest from the center and put it into $D_t$ to obtain $D_{t+1}$\;
    }
    \Else{
        \textbf{\texttt{break}}\;
    }
    \If {$|D_t|==|D_{t+1}|$}{
        \textbf{\texttt{break}}\;
    }
}
Construct the strong classifier $F(\bm{x})$ by Eq.\ref{eq16}\;
\textbf{return} $F(\bm{x})$.
\end{algorithm}

\subsection{Theoretical Justification}
\label{Justification}
In the context of the multiclass exponential loss function \cite{6}, the GAdaBoost.SA exhibits convergence, as demonstrated by the monotonic decrease of the loss function with each boosting iteration, provided that the base classifiers’ error rates remain below a certain threshold. This section briefly outlines the theoretical analysis process.

GAdaBoost.SA still uses the additive model in Eq.\ref{eq10}. Learning the additive model is equivalent to solving the problem of minimizing the empirical risk, that is, minimizing the loss function as Eq.\ref{eq11}.
The multiclass exponential loss function in Eq.\ref{eq1} is used to represent the loss of the $t$th iteration.
In the $(t+1)$th iteration, the strong classifier is $f_{t+1}(\bm{x}) = f_t(\bm{x}) + \beta_{t+1}h_{t+1}(\bm{x}; \gamma_{t+1})$. The error rate of the $(t+1)$th base classifier is denoted as Eq.\ref{eq19}.

\begin{equation}\label{eq19}
\epsilon_{t+1} = \frac{1}{|D_{t+1}|} \sum_{(\bm{x},y) \in D_{t+1}} \mathbb I(h_{t+1}(\bm{x};\gamma_{t+1}) \neq y).
\end{equation}
According to Eq.\ref{eq3}, the weight of the $(t+1)$th base classifier is $\beta_{t+1} = \log\left(\frac{1-\epsilon_{t+1}}{\epsilon_{t+1}}\right) + \log(K-1)$.

Then the loss function of the $(t+1)$th iteration is given in Eq.\ref{eq20}.
\begin{equation}
\begin{aligned}\label{eq20}
L(f_{t+1}(\bm{x})) &= \sum_{(\bm{x},y)\in D_{t+1}} \exp\left(-\frac{1}{K}\sum_{k=1}^K \mathbb I(y=k)(f_t^k(\bm{x}) + \beta_{t+1}\mathbb I(h_{t+1}(\bm{x};\gamma_{t+1})=k))\right)\\
&= \sum_{(\bm{x},y)\in D_{t+1}} \exp\left(-\frac{1}{K}\sum_{k=1}^K \mathbb I(y=k) f_t^k(\bm{x}) + -\frac{1}{K}\sum_{k=1}^K \mathbb I(y=k)\beta_{t+1}\mathbb I(h_{t+1}(\bm{x};\gamma_{t+1})=k)\right) \\
&= L(f_t(\bm{x})) \times \sum_{(\bm{x},y)\in D_{t+1}} \exp\left(-\frac{1}{K}\sum_{k=1}^K \mathbb I(y=k)\beta_{t+1}\mathbb I(h_{t+1}(\bm{x};\gamma_{t+1})=k)\right).
\end{aligned}
\end{equation}
Since the conditions of Eq.\ref{eq21} hold,
\begin{equation}\label{eq21}
\begin{cases}
  \mathbb I(y=k)=1, & \text{ if } y=k, \\
  \mathbb I(y=k)=0, & \text{ if } y \neq k.
\end{cases}
\end{equation}
Then, Eq.\ref{eq22} also holds.
\begin{equation}\label{eq22}
L(f_{t+1}(\bm{x})) = L(f_t(\bm{x})) \times \sum_{(\bm{x},y)\in D_{t+1}} \exp\left(-\frac{1}{K}\beta_{t+1}\mathbb I(h_{t+1}(\bm{x};\gamma_{t+1})=k)\right).
\end{equation}
For $\forall (\bm{x},y) \in D_{i+1}$, Eq.\ref{eq23} holds.
\begin{equation}\label{eq23}
\exp\left(-\frac{1}{K}\beta_{t+1}\mathbb I(h_{t+1}(\bm{x};\gamma_{t+1})=y)\right) = \begin{cases}
\exp(-\frac{1}{K}\beta_{t+1}), & \text{if } h_{t+1}(\bm{x};\gamma_{t+1})=y, \\
1, & \text{if } h_{t+1}(\bm{x};\gamma_{t+1}) \neq y.
\end{cases}
\end{equation}
Consider calculating the expected value of the loss function. The loss of the $(t+1)$th iteration can be expressed as Eq.\ref{eq24}.
\begin{equation}\label{eq24}
L(f_{t+1}) = L(f_t) \cdot \left[\epsilon_{t+1} + (1-\epsilon_{t+1})(\frac{\epsilon_{t+1}}{1-\epsilon_{t+1}})^\frac{1}{K}  \cdot (\frac{1}{K-1})^\frac{1}{K} \right].
\end{equation}
Let $R(\epsilon_{t+1}) = \epsilon_{t+1} + (1-\epsilon_{t+1})(\frac{\epsilon_{t+1}}{1-\epsilon_{t+1}})^\frac{1}{K} \cdot (\frac{1}{K-1})^\frac{1}{K}$, then $L(f_{t+1}) = L(f_t) \times R(\epsilon_{t+1})$. Let $u = \epsilon_{t+1}$, and take the derivative of $R(u)$ with respect to $u$.

\begin{equation}\label{eq25}
\begin{aligned}
\frac{d R(u)}{d u} &= 1 + \frac{d}{du}[(1-u)(\frac{u}{1-u})^\frac{1}{K} \cdot (\frac{1}{K-1})^\frac{1}{K}]\ \\
&= 1 + (\frac{1}{K-1})^\frac{1}{K}[\frac{d}{du}(1-u)(\frac{u}{1-u})^\frac{1}{K}]\ \\
&= 1 + (\frac{1}{K-1})^\frac{1}{K}[-(\frac{u}{1-u})^\frac{1}{K} + (1-u)\frac{1}{K}(\frac{u}{1-u})^{\frac{1}{K}-1}\frac{u+(1-u)}{(1-u)^2}]\ \\
&= 1 + (\frac{1}{K-1})^\frac{1}{K}(\frac{u}{1-u})^{\frac{1}{K}-1}[\frac{1-Ku}{K(1-u)}].
\end{aligned}
\end{equation}

Because $u \in [0,1)$, $K>=2$, the sign of $\frac{d R(u)}{du}$ can be inferred by examining $1-Ku$. On the interval $[0,\frac{1}{K}]$, $\frac{dR(u)}{du}>1$, and $R(u)$ is strictly monotonically increasing; on the interval $(\frac{1}{K},1)$, the monotonicity of $R(u)$ needs further analysis. Because $R(\frac{1}{K}) < 1$, $R(\frac{K-1}{K}) = 1$, $\frac{1}{K} < \frac{K-1}{K}$, and $R(u)$ is continuous on the $[0,1)$ interval. Then, for any $u \in [0,\frac{K-1}{K})$, $R(u) < 1$ must hold.
Therefore, the following inequality holds:
\begin{equation}\label{eq26}
L(f_{t+1}) < L(f_t).
\end{equation}

To summarize, the loss function in Eq.\ref{eq1} converges as the number of iterations increases.

\subsection{Computational Cost}
\label{Cost}
In Algorithm \ref{alg1}, assume that the entire granulation process iterates for a total of $M$ rounds, with $m_i$ IGs requiring splitting in each iteration. In the $i$th iteration, the division complexity of each IG is $O(|O_i^j| \cdot q)$, which arises from the following computational steps: (1) Computing the center vectors $\bm{c}_{i+1}^k $, requiring $ O(|O_i^j| \cdot q) $ operations; (2) Calculating the distance from each sample to all new centers, also requiring $ O(|O_i^j| \cdot q) $ operations; and (3) Assigning samples to new IGs, which takes at most $ O(|O_i^j|) $ operations. Thus, the total time complexity of the $i$th iteration is $O(m_i\cdot|O_i^j|\cdot q)$. Given that the total number of iterations $M$ remains bounded and is typically sublinear with respect to $N$, the overall worst-case time complexity is less than $ O(N \cdot q \cdot M) $.

In Algorithm \ref{alg2}, assume that a decision tree is used as the base classifier, the depth of the decision tree is $d$, and the data subset used to train the $t$-th base classifier is $D_t$. Then the training time complexity of the $t$th base classifier is $O(dq|D_i|\log|D_i|)$. Assume that the entire boosting process is iterated $T$ times. Because $D_t$ is dynamically updated, let the average size of each training subset be $\tau$. Then, the total training complexity of the base classifier is $O(T\cdot dq\cdot \tau \cdot\log(\tau))$. Assuming that unsampled samples are selected from $m^{\star}$ GBs in each iteration to update the training subset, the complexity is $O(T\cdot m^{\star})$. In summary, the total time complexity of Algorithm \ref{alg2} is $O(N\cdot q \cdot M+T(\tau \cdot dq\cdot\log(\tau )+m^{\star}))$, which is still linear. Notably, due to the early stopping mechanism of Algorithm \ref{alg2}, the actual iterations $T$ will be relatively small; that is, the time consumption will be relatively small.

\section{Experiments}
\label{experiments}
Detailed numerical experiments are conducted to verify the robustness, efficiency, and broad applicability of the proposed GAdaBoost.SA, with comparisons against the SAMME and its robust variant, Rob\_SAMME. All experiments are conducted on a Mac mini equipped with an Apple M4 chip and 24 GB of unified memory.

\subsection{Experimental Settings}
\label{Settings}
To quantify the performance of the proposed GAdaBoost.SA, the metrics $Accuracy$ and $F1-score$ are used. The total runtime (in ms) of each algorithm on each dataset is recorded to evaluate their efficiency. For visualization, the runtime is logarithmic. Each algorithm is repeated 5 times on each dataset, and the average value is used for comparison. For all datasets, a random $20\%$ of the samples is held out as a testing dataset. Notably, in the figures and tables of this section, GAdaBoost.SA, SAMME, and Rob\_SAMME are abbreviated as GSA, SA, and RSA, respectively.

\subsubsection{Dataset Description}
\label{Datasets}
Real-world datasets from the KEEL-dataset repository\cite{49} are utilized. These datasets include low-dimensional (e.g., svmguide1), high-dimensional (e.g., coil2000), small-scale (e.g., balance), large-scale (e.g., fars), binary (e.g., svmguide1), and multiclass types (e.g., penbased). In addition, several image classification datasets, such as MNIST, Fashion-MNIST, USPS, and OrganMNIST-Sagittal, are considered. MNIST is a classic handwritten digit image dataset that is widely used for benchmarking. Fashion-MNIST is a fashion product image dataset similar to MNIST, providing test scenarios with more complex patterns. USPS is a handwritten digit image dataset with fewer samples and lower image quality, making it more challenging. OrganMNIST-Sagittal is a medical image dataset containing organ slices from CT images for multi-class medical image classification tasks. Table \ref{table2} provides detailed information about these datasets.

To verify the robustness of GAdaBoost.SA to label noise, noisy datasets with label noise rates of $0.05, 0.1, 0.15, 0.2, 0.25$, and $0.3$ are constructed based on the above standard datasets. Specifically, for any standard dataset, a given proportion of samples is randomly selected, and their labels are replaced with other labels that are inconsistent with their true labels.

\begin{table}[H]
	\caption{Details of Datasets.}
	\label{table2}
	\centering
	\renewcommand\tabcolsep{1.5pt}
	\begin{tabular}{lccccc}
		\toprule[1pt]
		{Datasets} &{Rename} &{Samples} & {Features}& {Classes} &{Source}
        \\ \hline
    balance       &S1  &625     &4    &3     &\cite{49}  \\
    contraceptive      &S2  &1473     &9   &3     &\cite{49}  \\
    segment            &S3  &2310    &19   &7     &\cite{49} \\
    splice             &S4  &3190    &60   &3    &\cite{49}  \\
    page-blocks        &S5  &5473    &11   &5    &\cite{49} \\
    svmguide1           &S6  &6910    &4   &2    &\cite{49} \\
    thyroid             &S7  &7200    &21   &3     &\cite{49}  \\
    coil2000           &S8  &9822    &85   &2     &\cite{49} \\
    penbased          &S9  &10992   &16   &10     &\cite{49}      \\
    nursery            &S10  &12960   &8   &5     &\cite{49}      \\
    shuttle            &S11  &58000   &9   &7    &\cite{49} \\
    fars               &S12  &100968  &29  &8     &\cite{49}  \\
    USPS                &S13 &9298  &$16\times16$  &10    &\cite{52}  \\
    OrganMNIST-Sagittal &S14 &25221  &$28\times28$  &11    &\cite{51}  \\
    MNIST               &S15 &70000  &$28\times28$  &10    &\cite{53}  \\
    Fashion-MNIST       &S16 &70000  &$28\times28$  &10    &\cite{54}  \\
		\toprule[1pt]
	\end{tabular}
\end{table}

\subsubsection{Baselines and Parameter Settings}
\label{Baselines&Parameter}
GAdaBoost.SA is an extension of SAMME, and its main goal is to enhance the robustness of label noise scenarios while improving efficiency. Therefore, SAMME and Rob\_SAMME are used as baselines, both of which are introduced in detail in Section \ref{adaboost}. Furthermore, to demonstrate the broad applicability of GAdaBoost to various base classifiers, classification and regression trees (CART) is used in Sections \ref{KEEL} and \ref{Efficiency}, MLP is used in Section \ref{Image}, and support vector machines (SVM) is used in Section \ref{Ablation}.

SAMME and Rob\_SAMME are reproduced following the pseudocode in the corresponding paper, and CART and MLP are implemented using Scikit-Learn. To ensure fairness, the public parameters of GAdaBoost.SA and baselines, including the number of iterations and the depth of the CART, are tuned, whereas the other parameters are kept consistent with the default parameters in Scikit-Learn. The hyperparametric optimization framework Optuna (\cite{50}) is employed to determine fair parameters, with $100$ trials for each algorithm. In addition, five-fold cross-validation is used during the optimization process to obtain more stable and reliable performance estimates. The search space is designed considering the fast convergence of GAdaBoost.SA and the information provided in the corresponding paper. Specifically, $max\_depth \in[1,10]$ and $n\_estimators \in[10,200]$.

\subsection{Experiments on KEEL Datasets}
\label{KEEL}

\begin{figure}[H]
    \centering
    \subfigure[Noise rate: 5$\%$]{
        \includegraphics[height=5cm, width=5cm]{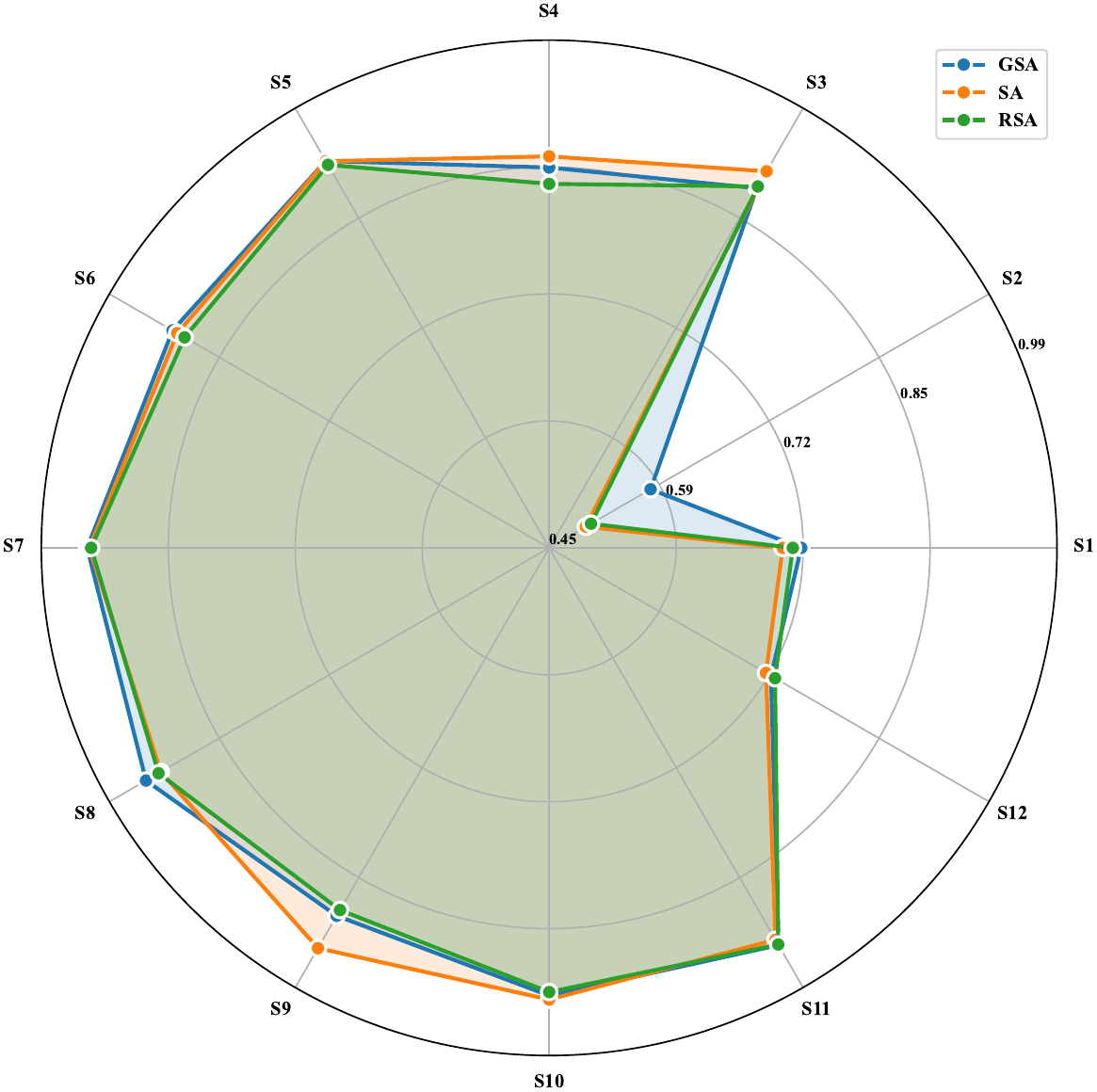}
    }
    \subfigure[Noise rate: 10$\%$]{
        \includegraphics[height=5cm, width=5cm]{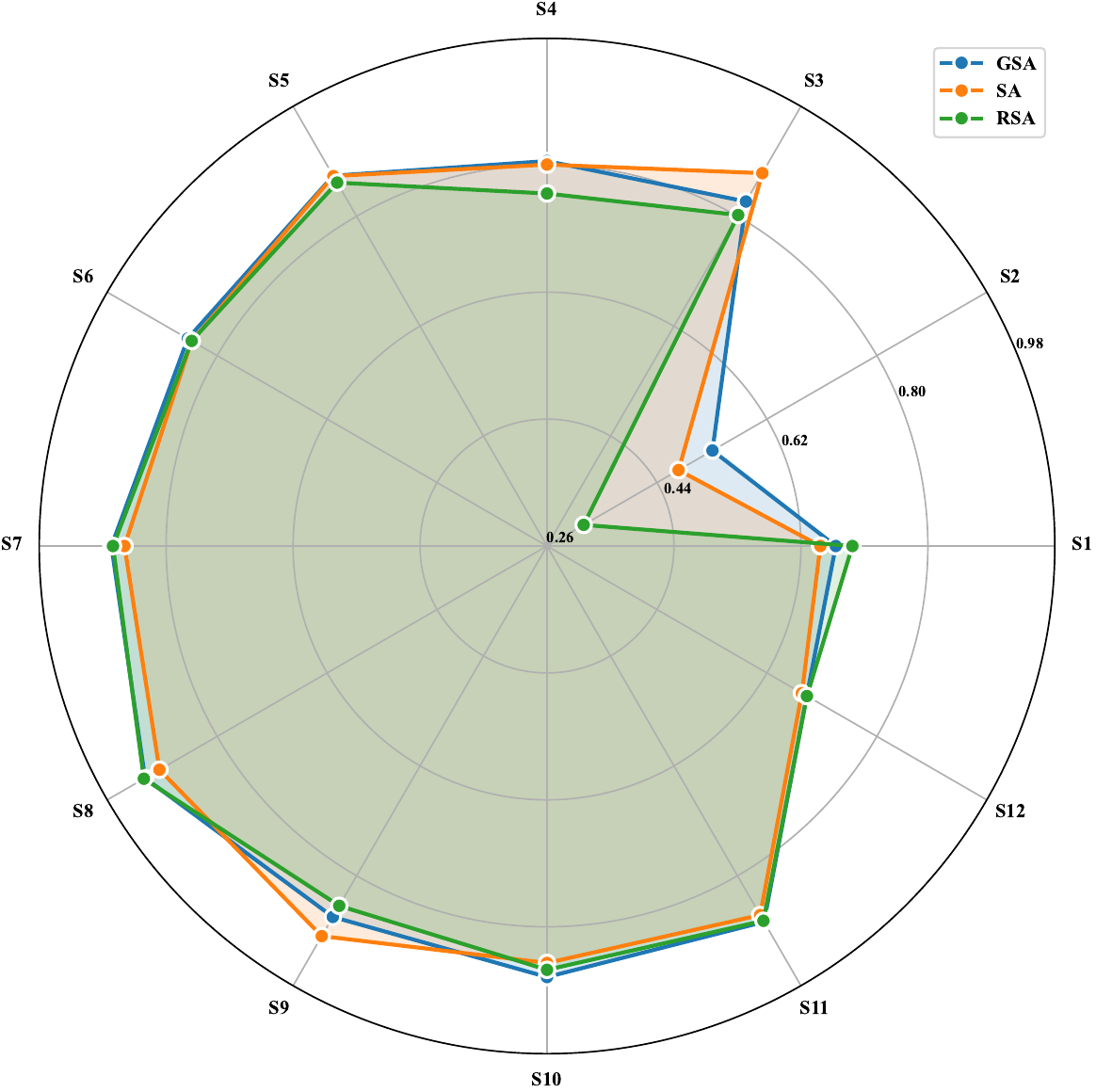}
    }
    \subfigure[Noise rate: 15$\%$]{
        \includegraphics[height=5cm, width=5cm]{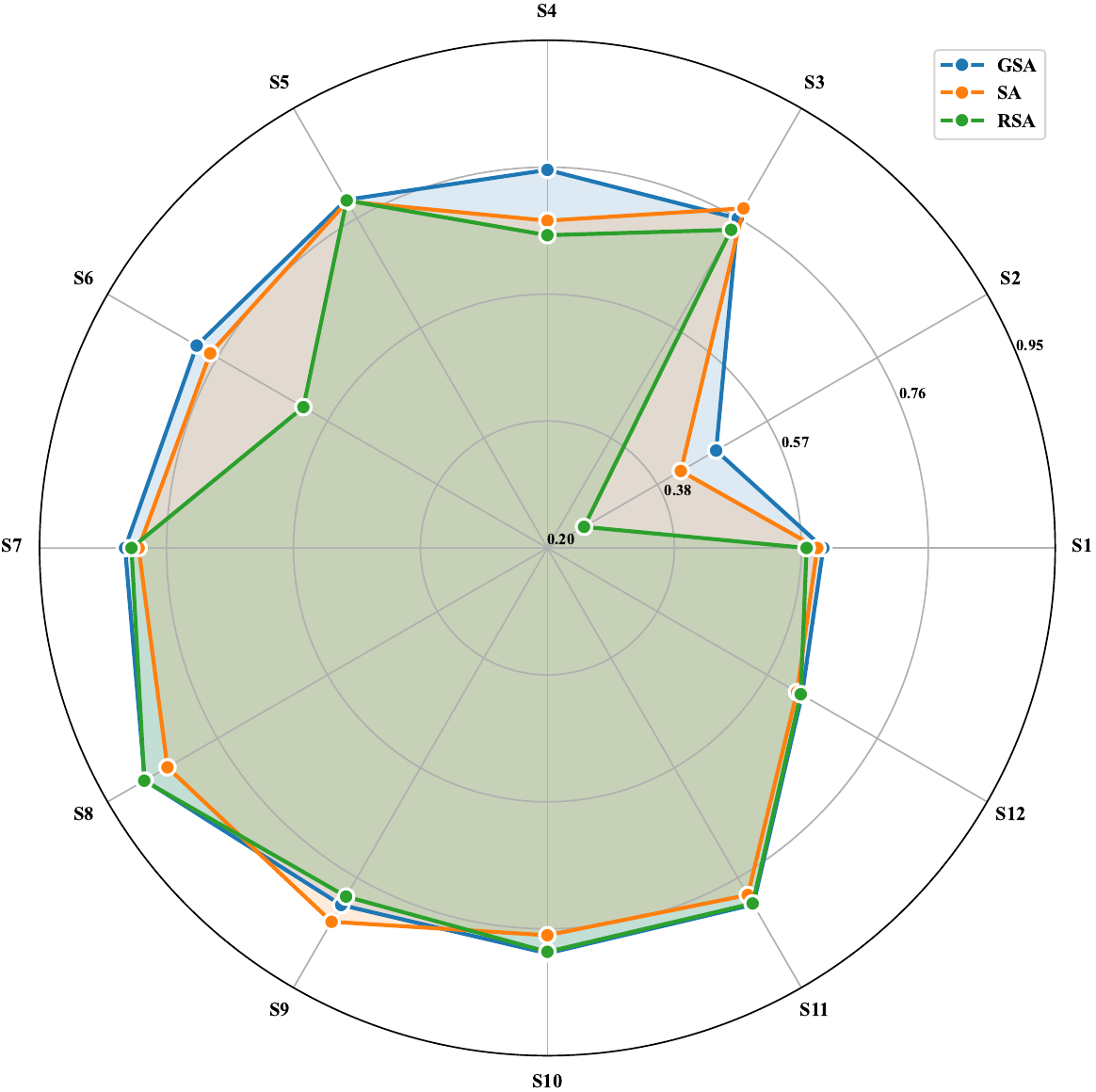}
    }\\ 
    \subfigure[Noise rate: 20$\%$]{
        \includegraphics[height=5cm, width=5cm]{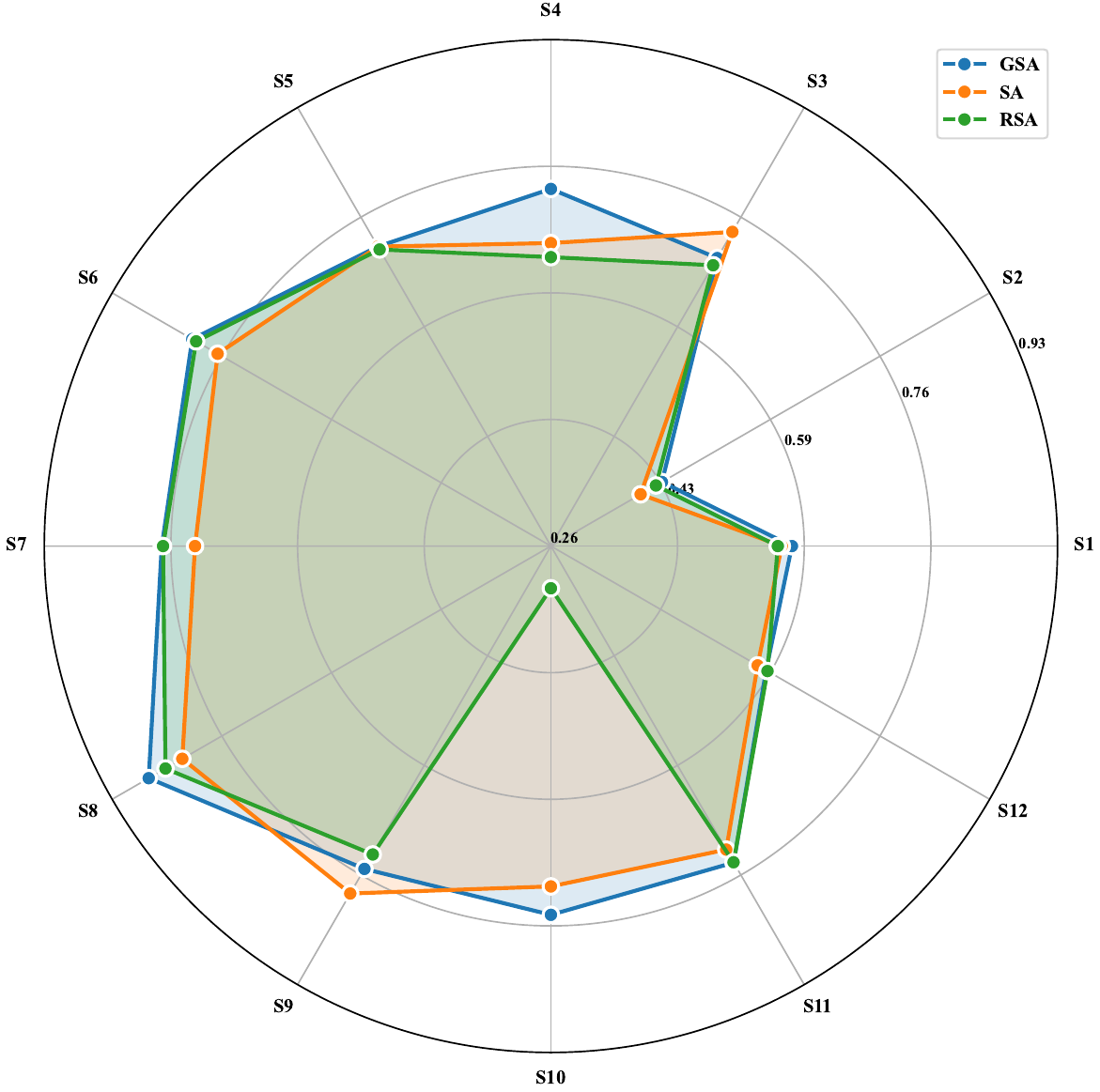}
    }
    \subfigure[Noise rate: 25$\%$]{
        \includegraphics[height=5cm, width=5cm]{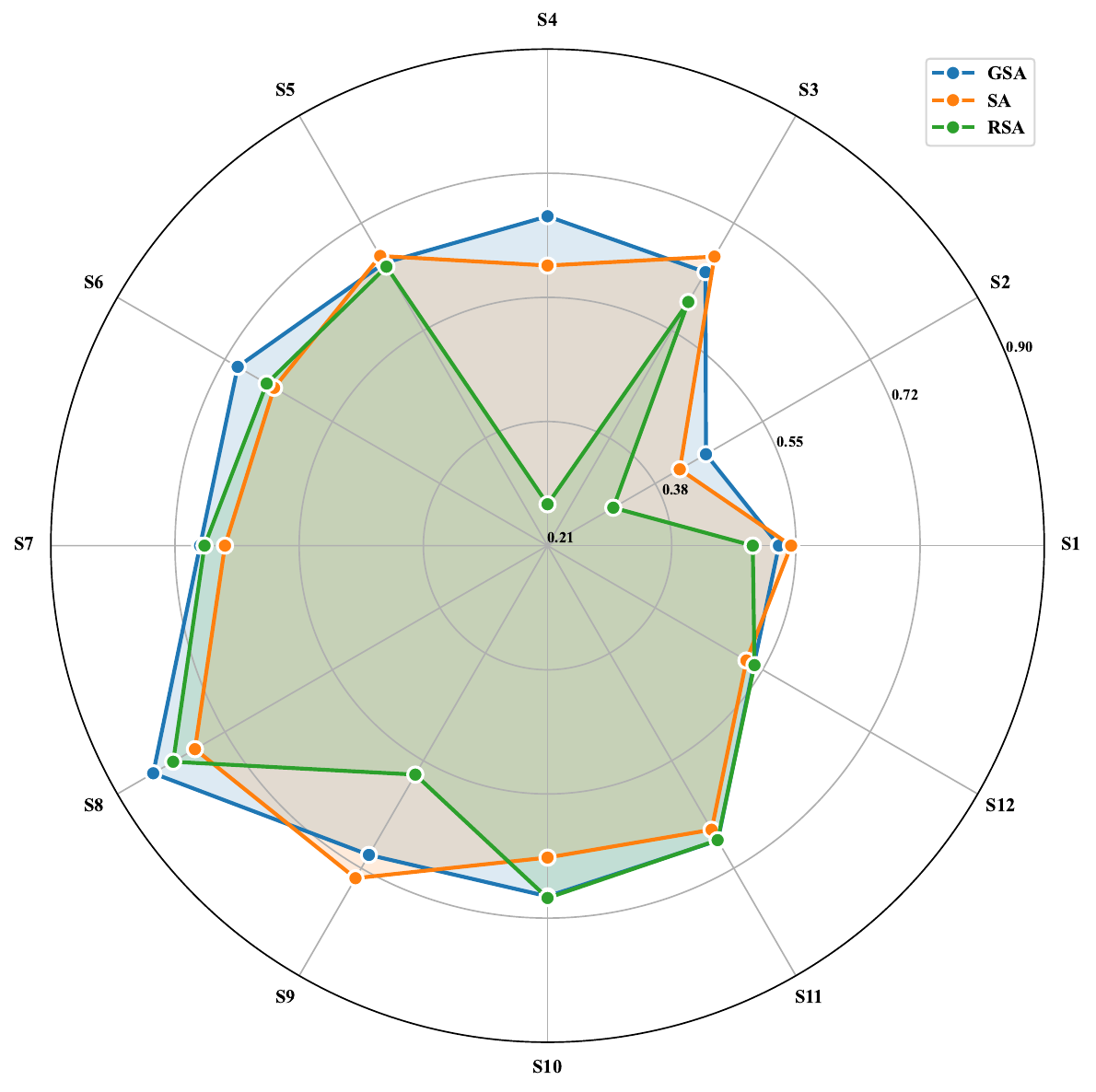}
    }
    \subfigure[Noise rate: 30$\%$]{
        \includegraphics[height=5cm, width=5cm]{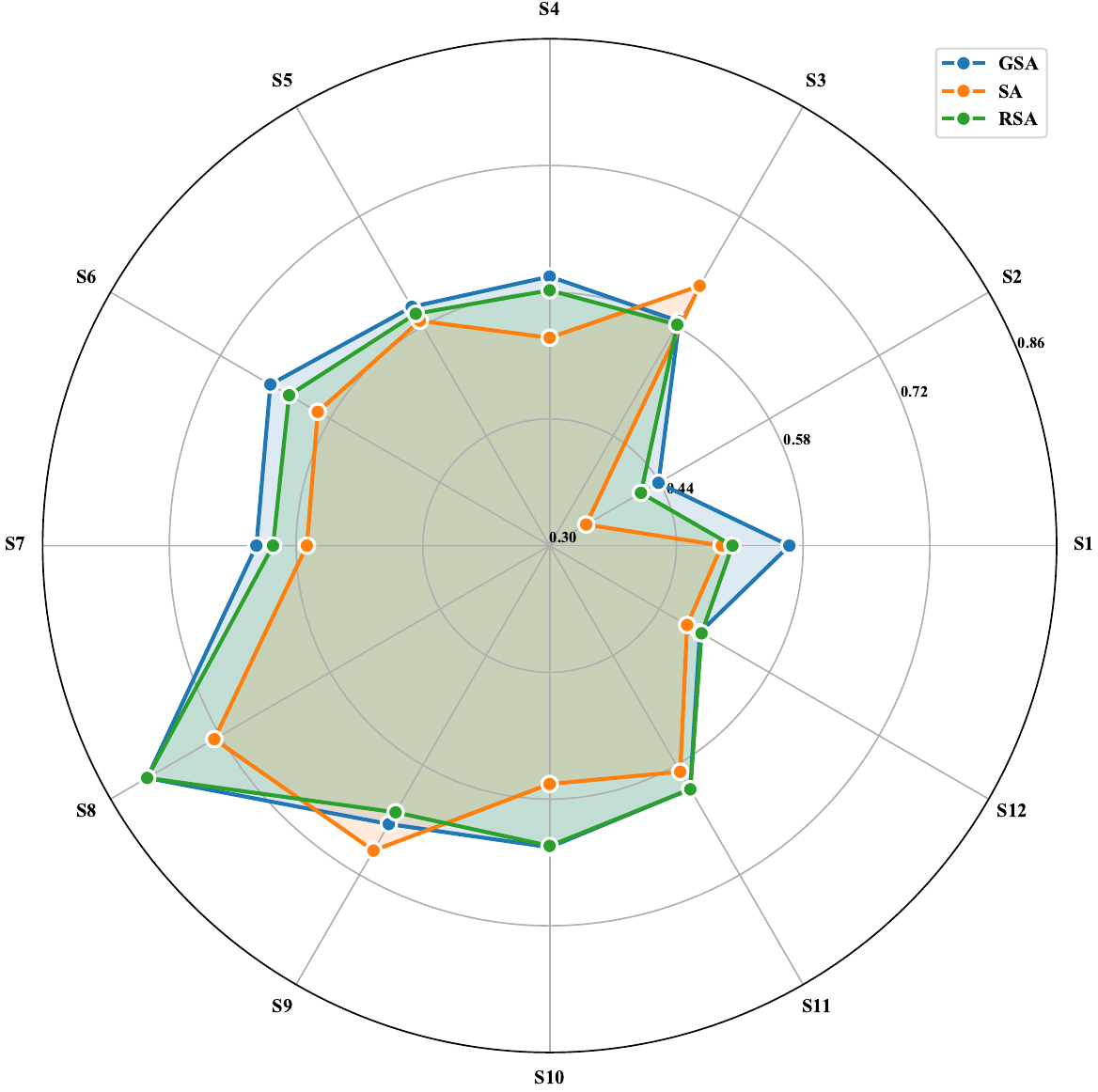}
    }
    \caption{Comparison of test $F1-score$ on KEEL datasets under different noise rates.}
    \label{fig7}
\end{figure}

This section presents experiments conducted on GAdaBoost.SA, Rob\_SAMME, and SAMME using 12 datasets from the KEEL repository at different label noise levels. The detailed test $Accuracy$ is shown in Tables \ref{table3} and \ref{table4}, with the best results highlighted in bold. Additionally, Figure \ref{fig7} presents a radar chart illustrating the comparative $F1-score$ performance of the three methods across all datasets and noise levels.

\begin{table}[H]
	\caption{Comparison of test $Accuracy$ on KEEL datasets. (Noise rate from 5$\%$ to 15$\%$)}
	\label{table3}
	\centering
	\renewcommand\tabcolsep{1.5pt}
        \scriptsize  
        \resizebox{\textwidth}{!}{
        \begin{tabular}{p{1.5cm}p{1.3cm}p{1.3cm}p{1.3cm}p{1.3cm}p{1.3cm}p{1.3cm}p{1.3cm}p{1.3cm}p{1.3cm}}
	\toprule[1pt]
\multirow{2}{*}{Dataset}	&\multicolumn{3}{c}{Noise rate: 5$\%$} & \multicolumn{3}{c}{Noise rate: 10$\%$} &\multicolumn{3}{c}{Noise rate: 15$\%$}
\\ \cmidrule(lr){2-4} \cmidrule(lr){5-7} \cmidrule(lr){8-10}
	 &GSA &SA &RSA	&GSA &SA &RSA 	&GSA &SA &RSA
\\ \hline
    S1    &\textbf{0.76}    &0.696   &0.728    &0.688  &0.64    &\textbf{0.752}   &\textbf{0.664}   &0.6     &0.64     \\
    S2    &\textbf{0.5763}  &0.5288  &0.539    &\textbf{0.5458} &0.5186  &0.5322  &\textbf{0.4949}  &\textbf{0.4949}  &0.4305   \\
    S3    &0.8961  &\textbf{0.9156}  &0.8918   &0.8333 &\textbf{0.868}   &0.816   &0.7749  &\textbf{0.7792}  &0.7424   \\
    S4    &0.8652  &\textbf{0.873}   &0.8433   &\textbf{0.8135} &0.7947  &0.8088  &\textbf{0.7555}  &0.7524  &0.7069   \\
    S5    &\textbf{0.9323}  &0.9314  &0.9268   &\textbf{0.8851} &0.8814  &0.8786  &\textbf{0.8304}  &0.8285  &0.8258   \\
    S6    &\textbf{0.9276}  &0.9226  &0.9139   &\textbf{0.8669} &0.864   &0.8603  &\textbf{0.8256}  &0.8046  &0.4436   \\
    S7    &\textbf{0.9444}  &0.941   &0.9403   &\textbf{0.8931} &0.8833  &0.8917  &\textbf{0.8528}  &0.85    &0.8424   \\
    S8    &\textbf{0.8917}  &\textbf{0.8917}  &\textbf{0.8917}   &0.8439 &0.8433  &\textbf{0.8451}  &0.7961  &0.7955  &\textbf{0.7973}   \\
    S9    &0.9     &\textbf{0.9395}  &0.8927   &0.8627 &\textbf{0.895}   &0.8454  &0.8081  &\textbf{0.8372}  &0.7954   \\
    S10   &0.9309  &\textbf{0.9313}  &0.9263   &\textbf{0.8827} &0.8588  &0.8723  &\textbf{0.8206}  &0.7998  &0.814    \\
    S11   &\textbf{0.9494}  &0.9412  &0.9484   &\textbf{0.9003} &0.8851  &0.8999  &\textbf{0.8505}  &\textbf{0.8505}  &0.8483   \\
    S12   &0.7573  &0.7544  &\textbf{0.7598}   &\textbf{0.7235} &0.7225  &0.7229  &\textbf{0.6817}  &0.6814  &0.6813   \\
\toprule[1pt]
\end{tabular}
}
\end{table}

As shown in Tables \ref{table3} and \ref{table4}, GAdaBoost.SA outperforms Rob\_SAMME and SAMME on most datasets across all noise rates, and this advantage becomes more pronounced as the noise rate increases, as confirmed by the Wilcoxon signed-rank test and Win/Loss/Tie Analysis (Table \ref{table5}). Although the performance differences are not always statistically significant, GAdaBoost.SA consistently achieves more wins than the two baselines, demonstrating its robustness under varying noise conditions. This is mainly attributed to the GB structure, which removes part of the noisy samples while focusing on key class boundary samples within the GB. Combined with its incremental sampling strategy, GAdaBoost.SA effectively mitigates the influence of noise by guiding subsequent base classifiers to focus on challenging regions. In conclusion, the GBG-Ens method in GAdaBoost.SA adapts to dataset characteristics and effectively handles label noise. By contrast, SAMME assigns increasing weights to all misclassified samples, making it susceptible to label noise. Meanwhile, the performance of Rob\_SAMME heavily depends on the choice of $k$ in $k$NN, leading to instability across different datasets.

\begin{table}[H]
	\caption{Comparison of test $Accuracy$ on KEEL datasets. (Noise rate from 20$\%$ to 30$\%$)}
	\label{table4}
	\centering
	\renewcommand\tabcolsep{1.5pt}
        \scriptsize  
       \resizebox{\textwidth}{!}{
       \begin{tabular}{p{1.5cm}p{1.3cm}p{1.3cm}p{1.3cm}p{1.3cm}p{1.3cm}p{1.3cm}p{1.3cm}p{1.3cm}p{1.3cm}}
	\toprule[1pt]
\multirow{2}{*}{Dataset}	&\multicolumn{3}{c}{Noise rate: 20$\%$} & \multicolumn{3}{c}{Noise rate: 25$\%$} &\multicolumn{3}{c}{Noise rate: 30$\%$}
\\ \cmidrule(lr){2-4} \cmidrule(lr){5-7} \cmidrule(lr){8-10}
	 &GSA &SA &RSA	&GSA &SA &RSA 	&GSA &SA &RSA
\\ \hline
    S1    &\textbf{0.624}     &0.56    &0.552     &\textbf{0.576}     &0.56    &0.56     &\textbf{0.584}  &0.544   &0.504   \\
    S2    &\textbf{0.4441}    &0.4203  &0.4237    &\textbf{0.4847}    &0.4203  &0.4271   &\textbf{0.4407} &0.4237  &0.4203  \\
    S3    &0.7129    &\textbf{0.7359}  &0.6948    &0.6494    &\textbf{0.6688}  &0.6537   &0.5974 &\textbf{0.6407}  &0.5844  \\
    S4    &\textbf{0.7304}    &0.6614  &0.6379    &\textbf{0.6787}    &0.6317  &0.674    &\textbf{0.6113} &0.5705  &0.5831  \\
    S5    &\textbf{0.7757}    &0.7692  &0.7618    &\textbf{0.7349}    &0.7294  &0.7229   &\textbf{0.6886} &0.6766  &0.6728  \\
    S6    &\textbf{0.7815}    &0.7793  &0.7764    &\textbf{0.7352}    &0.7221  &0.7344   &\textbf{0.6896} &0.6527  &0.67    \\
    S7    &\textbf{0.8118}    &0.8111  &0.8111    &\textbf{0.7549}    &0.7035  &0.7361   &\textbf{0.6972} &0.6667  &0.6646  \\
    S8    &\textbf{0.7713}    &0.7683  &0.7695    &\textbf{0.7241}    &0.7042  &0.7217   &\textbf{0.6836} &0.6751  &\textbf{0.6836}  \\
    S9    &0.7503    &\textbf{0.7885}  &0.7281    &0.7035    &\textbf{0.7408}  &0.6871   &0.6558 &\textbf{0.688}   &0.6407  \\
    S10   &\textbf{0.7728}    &0.7218  &0.7681    &\textbf{0.7292}    &0.7203  &0.7249   &0.6752 &0.6613  &\textbf{0.6775}  \\
    S11   &\textbf{0.7994}    &0.7991  &0.7978    &\textbf{0.7493}    &0.7104  &0.7489   &\textbf{0.6951} &0.6833  &0.6942  \\
    S12   &0.644     &0.6412  &\textbf{0.6455}    &\textbf{0.6018}    &0.5911  &0.6006   &\textbf{0.5639} &0.5408  &0.5609  \\
\toprule[1pt]
\end{tabular}
}
\end{table}

\begin{table}[htbp]
	\renewcommand{\arraystretch}{1.0}
    \captionsetup{width=0.73\textwidth, justification=centering}
	\caption{Wilcoxon Signed-Rank Test and Win/Loss/Tie Analysis for $Accuracy$. ($\uparrow$ indicates statistical significance)}
	\label{table5}
	\centering
    \footnotesize
	\renewcommand\tabcolsep{3.5pt}
\begin{tabular}{ccccc}
	\toprule[1pt]
\multirow{2}{*}{Noise rates}	&\multicolumn{2}{c}{GSA vs. SA} &\multicolumn{2}{c}{GSA vs. RSA}
\\ \cmidrule(lr){2-3} \cmidrule(lr){4-5}
	 &p-value &Win/Lose/Tie	&p-value &Win/Lose/Tie
\\ \hline
$5\%$        &0.4236 &7/4/1   &0.0058$\uparrow$ &10/1/1 \\
$10\%$        &0.1763 &10/2/0   &0.0640$\uparrow$ &10/2/0 \\
$15\%$        &0.2026 &8/2/2   &0.0015$\uparrow$ &11/1/0 \\
$20\%$        &0.0771$\uparrow$ &10/2/0   &0.0015$\uparrow$ &11/1/0 \\
$25\%$        &0.0522$\uparrow$ &10/2/0   &0.0068$\uparrow$ &11/1/0 \\
$30\%$        &0.1514 &10/2/0   &0.0058$\uparrow$ &10/1/1\\
\toprule[1pt]
\end{tabular}
\end{table}

However, GAdaBoost.SA does not always achieve the best results.  For example, on the splice dataset with a $5\%$ noise rate, GAdaBoost.SA underperforms SAMME. This is because, for high-dimensional and small datasets (e.g., splice), the sample size of each GB is relatively small, and GAdaBoost.SA may include too many samples when constructing the initial training subset, leading to insufficient diversity in the base classifiers. Similarly, for datasets with many classes but sparse samples (e.g., penbased), GAdaBoost.SA may struggle to accurately capture class boundaries during the construction of the training subset, resulting in worse performance compared to SAMME, which trains on the entire dataset.

As illustrated in Figure \ref{fig7}, the radar chart reveals that GAdaBoost.SA consistently achieves a higher $F1-score$ than both SAMME and Rob\_SAMME across most datasets and noise levels, with particularly notable improvements over Rob\_SAMME. The broader and more stable shape of GAdaBoost.SA’s radar plot confirms its robustness, and this performance gap becomes even more pronounced as noise increases. This suggests that GAdaBoost.SA can better resist noisy labels through its granular adaptive boosting mechanism, maintaining superior classification balance and generalization even under high-noise scenarios.

\begin{table}[H]
	\caption{Optimal parameter settings for $Accuracy$ on each KEEL dataset.}
	\label{table6}
	\centering
	\renewcommand\tabcolsep{1.0pt}
        \scriptsize  
        \resizebox{\textwidth}{!}{
        \begin{tabular}{ccccccccccccc}
	\toprule[1pt]
\multirow{2}{*}{Dataset} &\multicolumn{2}{c}{Noise rate:5$\%$} & \multicolumn{2}{c}{Noise rate:10$\%$} &\multicolumn{2}{c}{Noise rate:15$\%$} &\multicolumn{2}{c}{Noise rate:20$\%$} & \multicolumn{2}{c}{Noise rate:25$\%$} &\multicolumn{2}{c}{Noise rate:30$\%$}
\\ \cmidrule(lr){2-3} \cmidrule(lr){4-5} \cmidrule(lr){6-7} \cmidrule(lr){8-9} \cmidrule(lr){10-11} \cmidrule(lr){12-13}
	 &depth &Iterations  &depth &Iterations &depth &Iterations &depth &Iterations   &depth &Iterations &depth &Iterations
        \\ \hline
        S1   &1 &86 &7 &11 &5 &10 &5 &13 &5 &70 &6 &10      \\
        S2   &7 &156 &4 &64 &3 &32 &3 &10 &3 &15 &5 &33     \\
        S3   &6 &82 &10 &11 &7 &22 &7 &40 &6 &20 &6 &17     \\
        S4   &6 &183 &5 &20 &5 &30 &5 &69 &5 &10 &6 &15     \\
        S5   &7 &186 &5 &35 &4 &10 &5 &32 &4 &23 &4 &23     \\
        S6   &3 &113 &4 &18 &4 &14 &4 &32 &5 &10 &5 &11     \\
        S7   &9 &30 &5 &11 &4 &63 &4 &10 &3 &54 &3 &10      \\
        S8   &2 &131 &2 &10 &2 &10 &1 &12 &1 &10 &1 &12     \\
        S9   &10 &116 &10 &18 &10 &43 &9 &69 &10 &31 &9 &12 \\
        S10   &7 &123 &10 &38 &9 &20 &9 &48 &9 &10 &8 &34    \\
        S11   &3 &93 &5 &16 &6 &10 &6 &54 &7 &25 &6 &10      \\
        S12   &10 &178 &10 &10 &8 &51 &8 &31 &8 &14 &9 &83   \\
\toprule[1pt]
\end{tabular}
}
\end{table}

Finally, Table \ref{table6} reports the hyperparameters searched by Optuna for optimal $Accuracy$ at different noise levels. The results reveal that GAdaBoost.SA tends to select shallow trees (1-5 levels) for simple datasets (e.g., S6 and S8), medium-depth trees (4-7 levels) for moderately complex datasets (e.g., S5 and S11), and deeper trees (7-10 levels) for high-dimensional or complex datasets (e.g., S3, S4, and S12). This reflects GAdaBoost.SA's ability to adaptively balance model complexity and generalization.

\subsection{Efficiency Analysis}
\label{Efficiency}

\begin{figure}[H]
    \centering
    \subfigure[Noise rate: 5$\%$]{
        \includegraphics[width=0.45\textwidth]{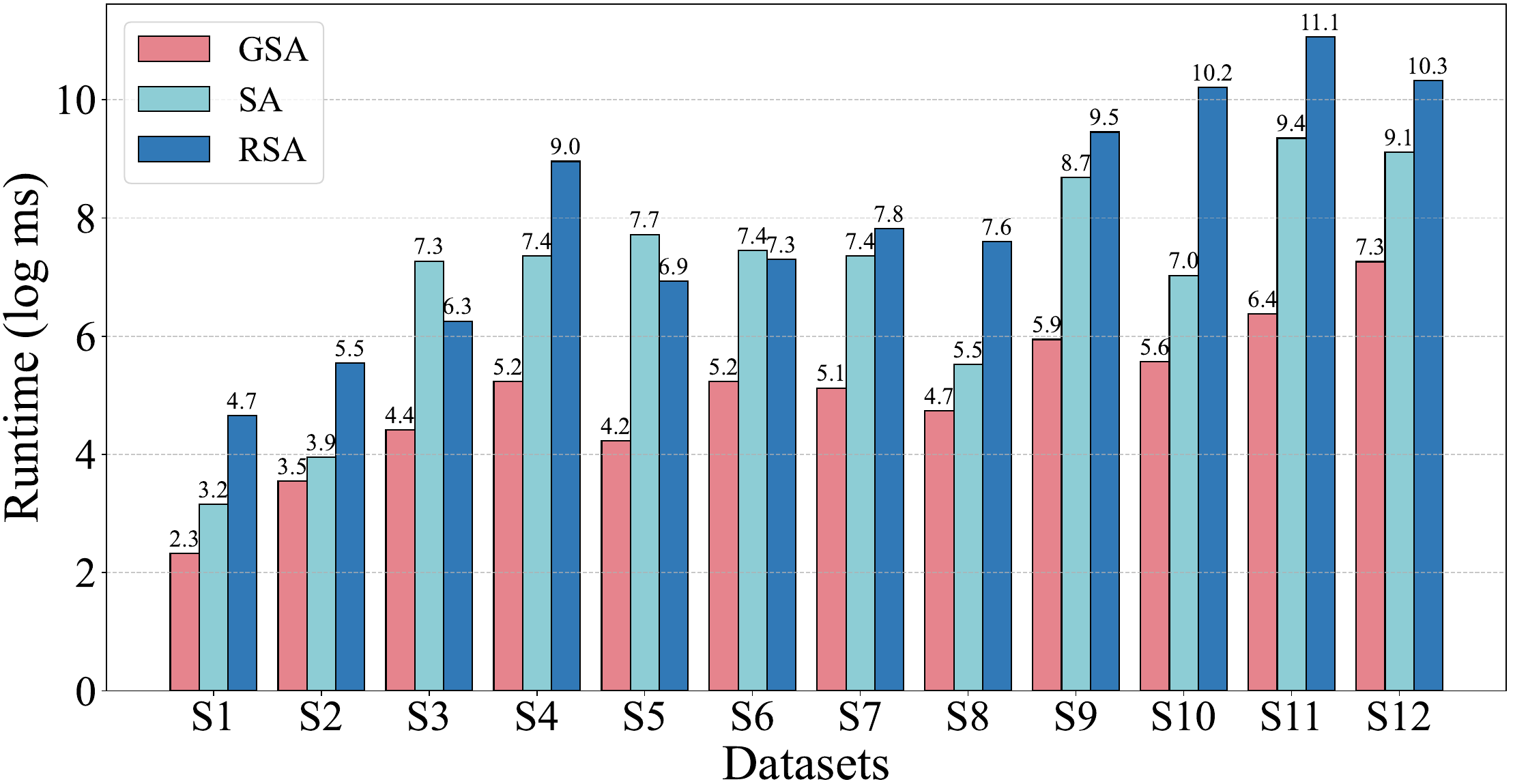}
    }
    \subfigure[Noise rate: 10$\%$]{
        \includegraphics[width=0.45\textwidth]{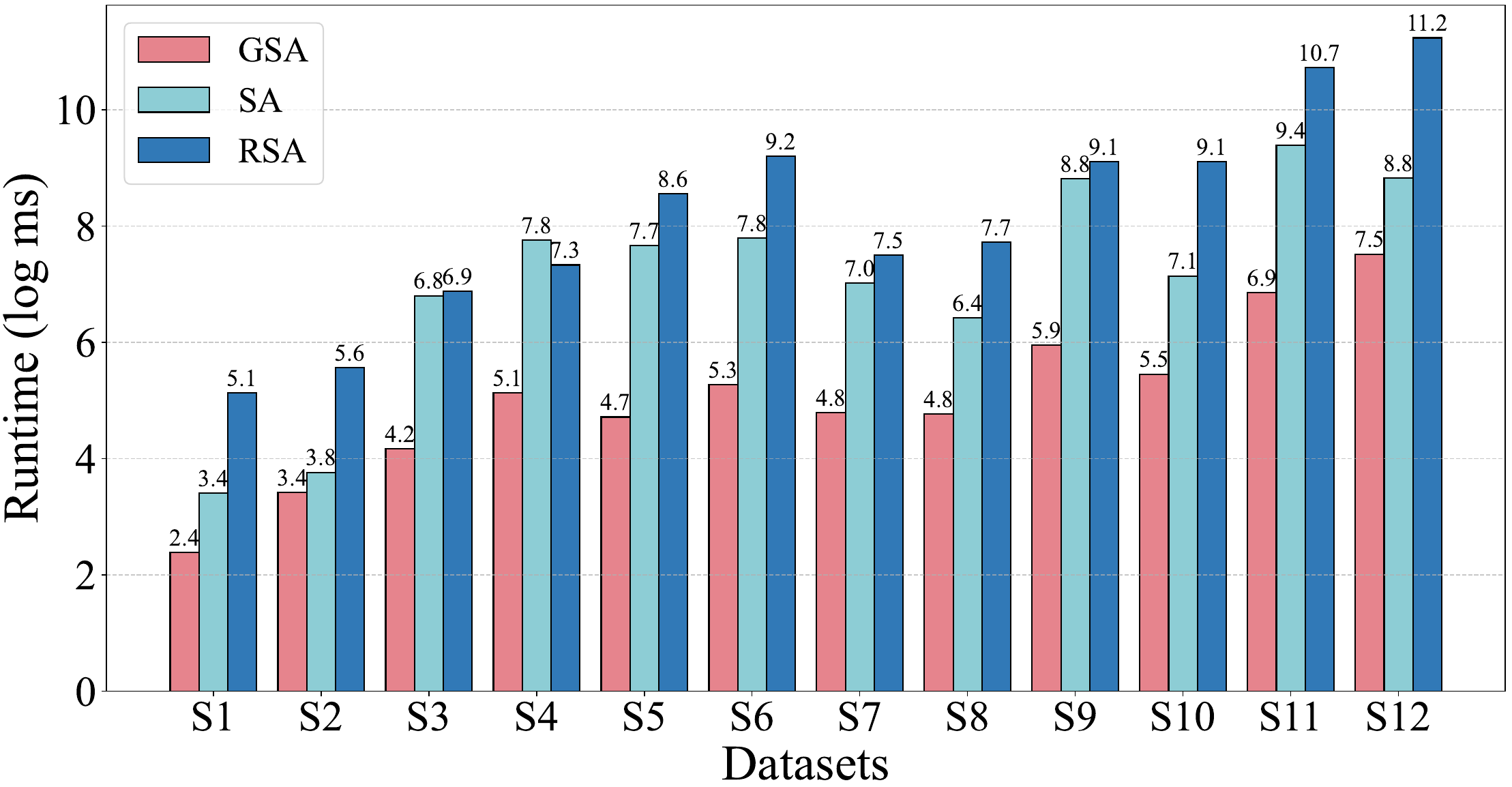}
    }
    \subfigure[Noise rate: 15$\%$]{
        \includegraphics[width=0.45\textwidth]{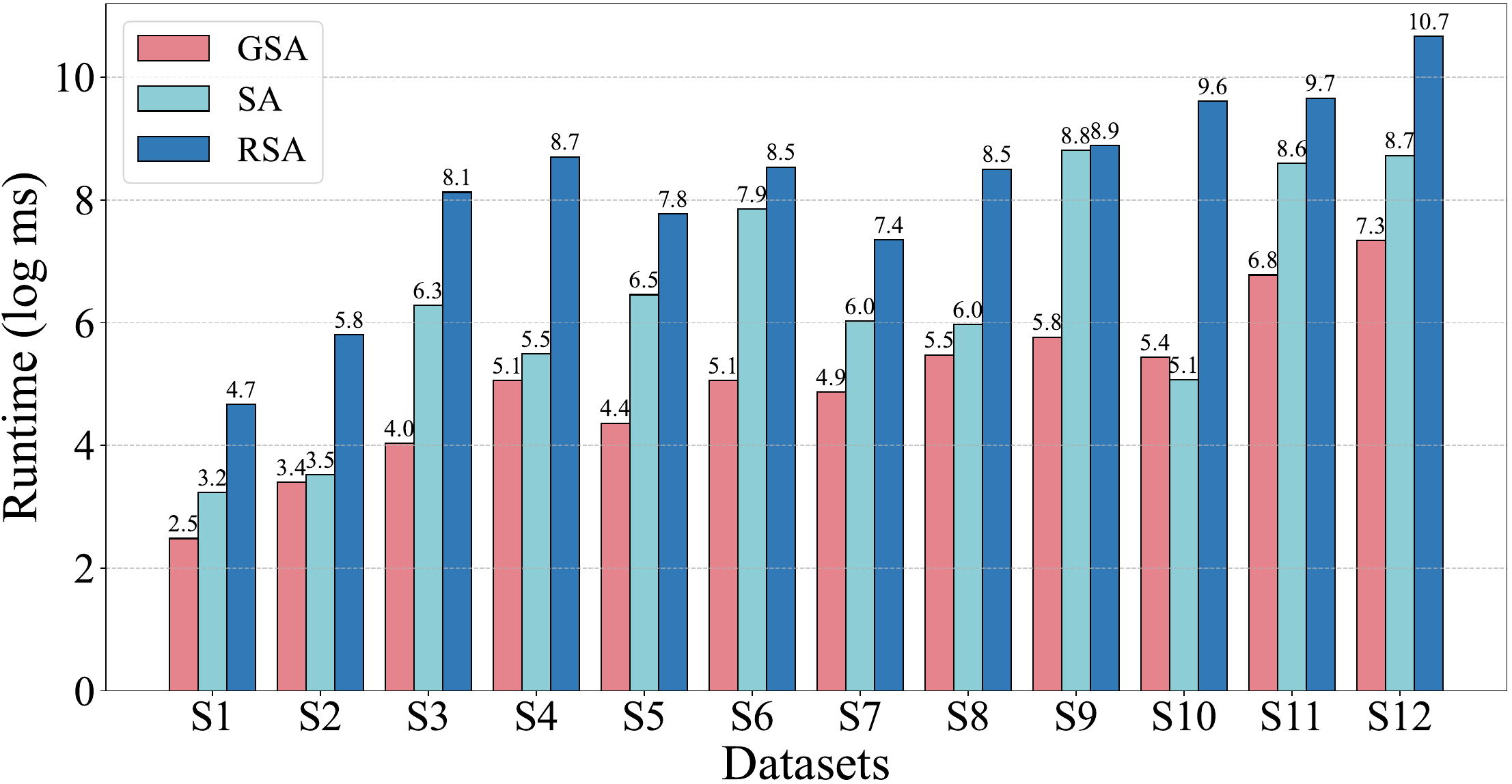}
    }
    \subfigure[Noise rate: 20$\%$]{
        \includegraphics[width=0.45\textwidth]{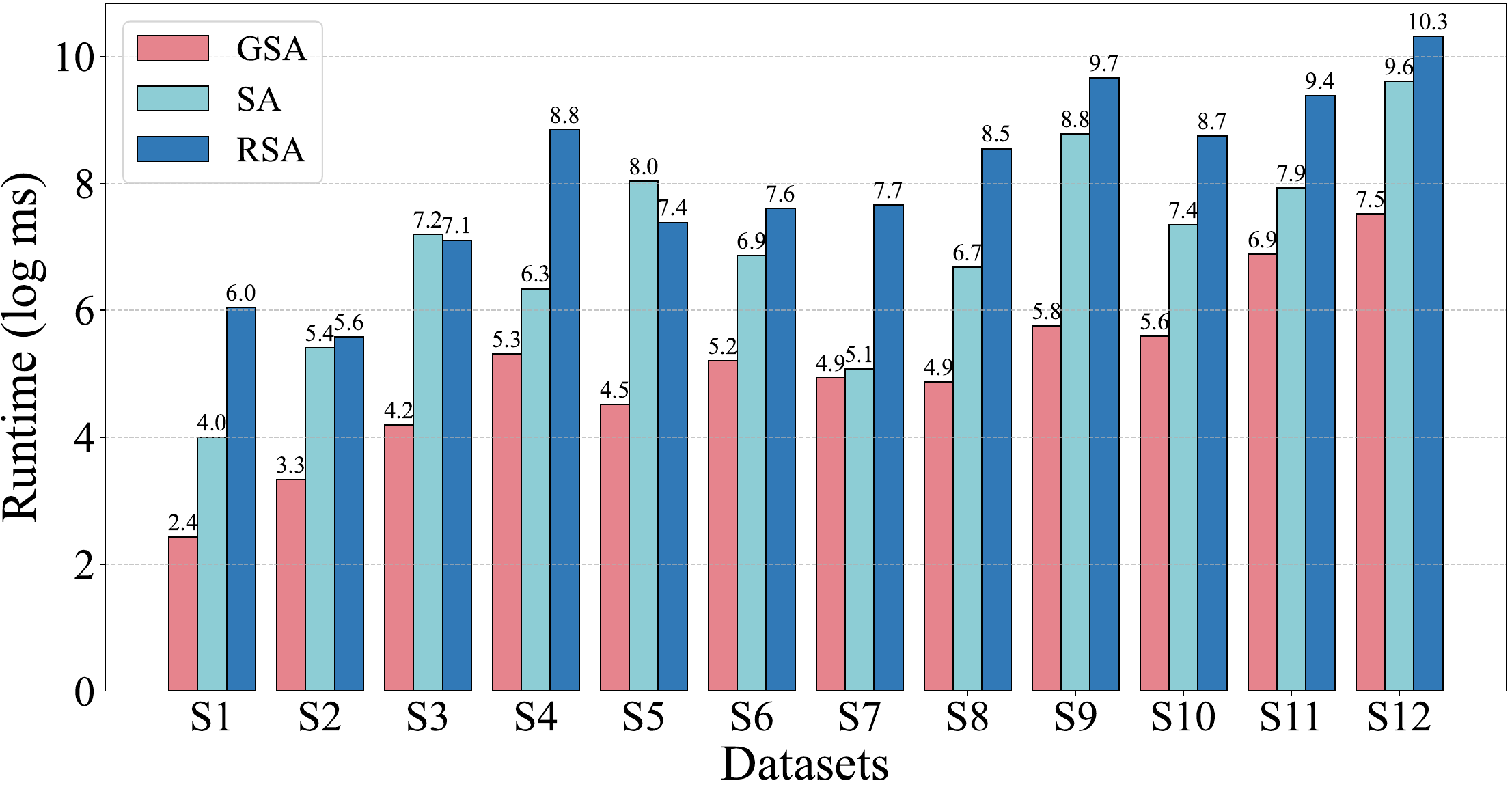}
    }
    \subfigure[Noise rate: 25$\%$]{
        \includegraphics[width=0.45\textwidth]{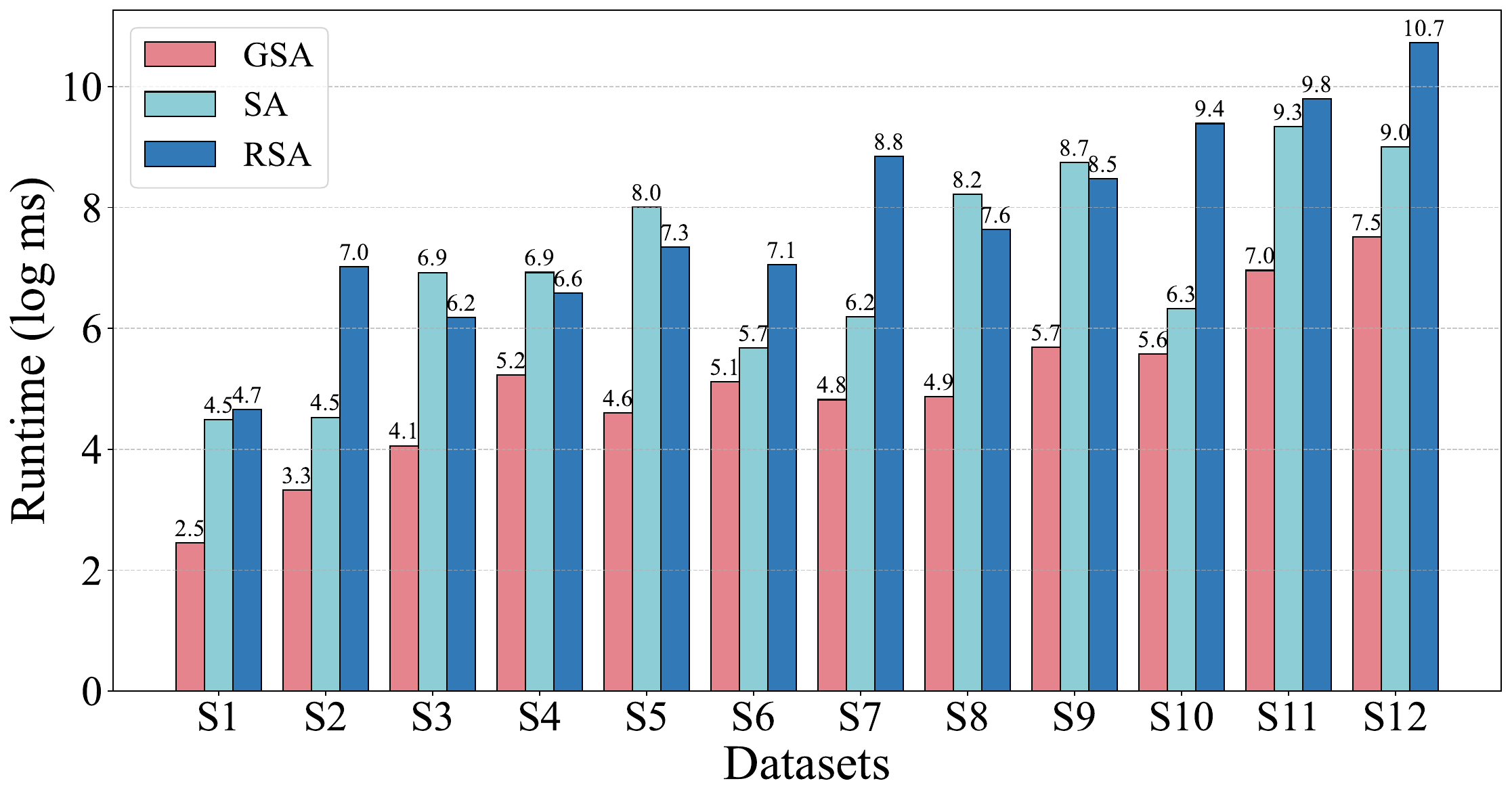}
    }
    \subfigure[Noise rate: 30$\%$]{
        \includegraphics[width=0.45\textwidth]{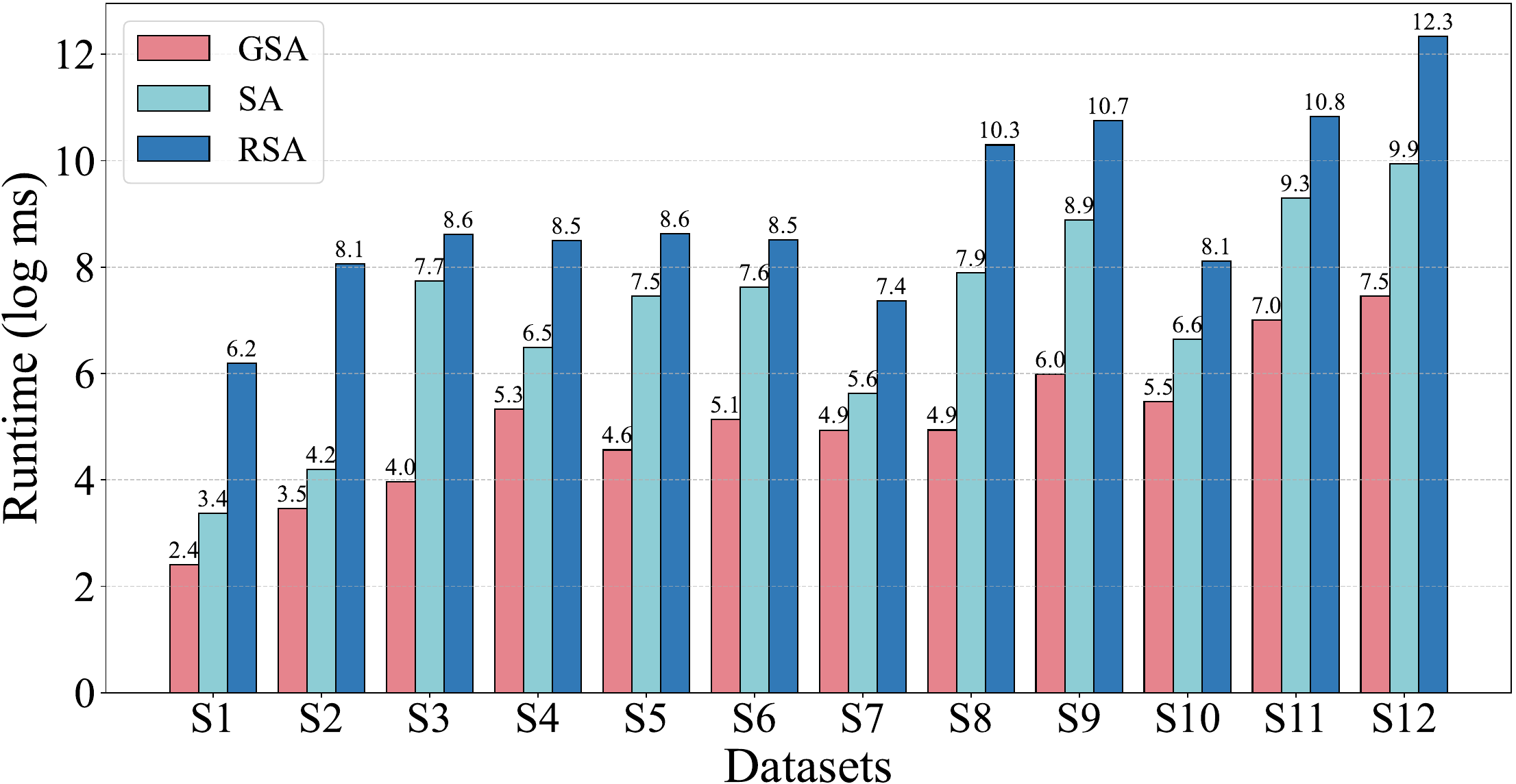}
    }
    \caption{Comparison of total runtime on KEEL datasets under different noise rates.}
    \label{fig4}
\end{figure}

Figure \ref{fig4} shows the total runtime of GAdaBoost.SA, Rob\_SAMME, and SAMME on 12 KEEL datasets under different noise levels. It can be seen that GAdaBoost.SA achieves the best efficiency in all cases. Compared with SAMME, the acceleration effect of GAdaBoost.SA is between several times and dozens of times. Compared with Rob\_SAMME, the acceleration effect of GAdaBoost.SA is even greater, ranging from several times to hundreds of times.

The reasons why GAdaBoost.SA achieves such a significant acceleration are as follows. First, GAdaBoost.SA granulates the training dataset by constructing GBs. Then, each iteration only needs to implicitly weigh the GBs, unlike SAMME and Rob\_SAMME, which need to weigh each sample individually, greatly reducing the number of objects that need to be weighed. Second, GAdaBoost.SA adopts an incremental sampling strategy, and each iteration only focuses on the GBs containing misclassified samples in the previous iteration and adds a small number of new samples to the subset of the subsequent base classifier. This avoids the redundant calculation of the entire dataset used by SAMME and Rob\_SAMME in each iteration, thus significantly improving efficiency. Third, GAdaBoost.SA has an adaptive early-stopping mechanism, where the iteration terminates early when the training subset converges or the error rate of the base classifier is zero. Fourth, GAdaBoost.SA identifies noise samples before inputting the data into the base classifier, whereas Rob\_SAMME needs to process noise samples using $k$NN in each iteration, which significantly increases the training time.

\subsection{Experiments on Image Datasets}
\label{Image}
\begin{figure}[H]
    \centering
    \subfigure[Fashion-MNIST (S16)]{
        \includegraphics[width=0.4\textwidth]{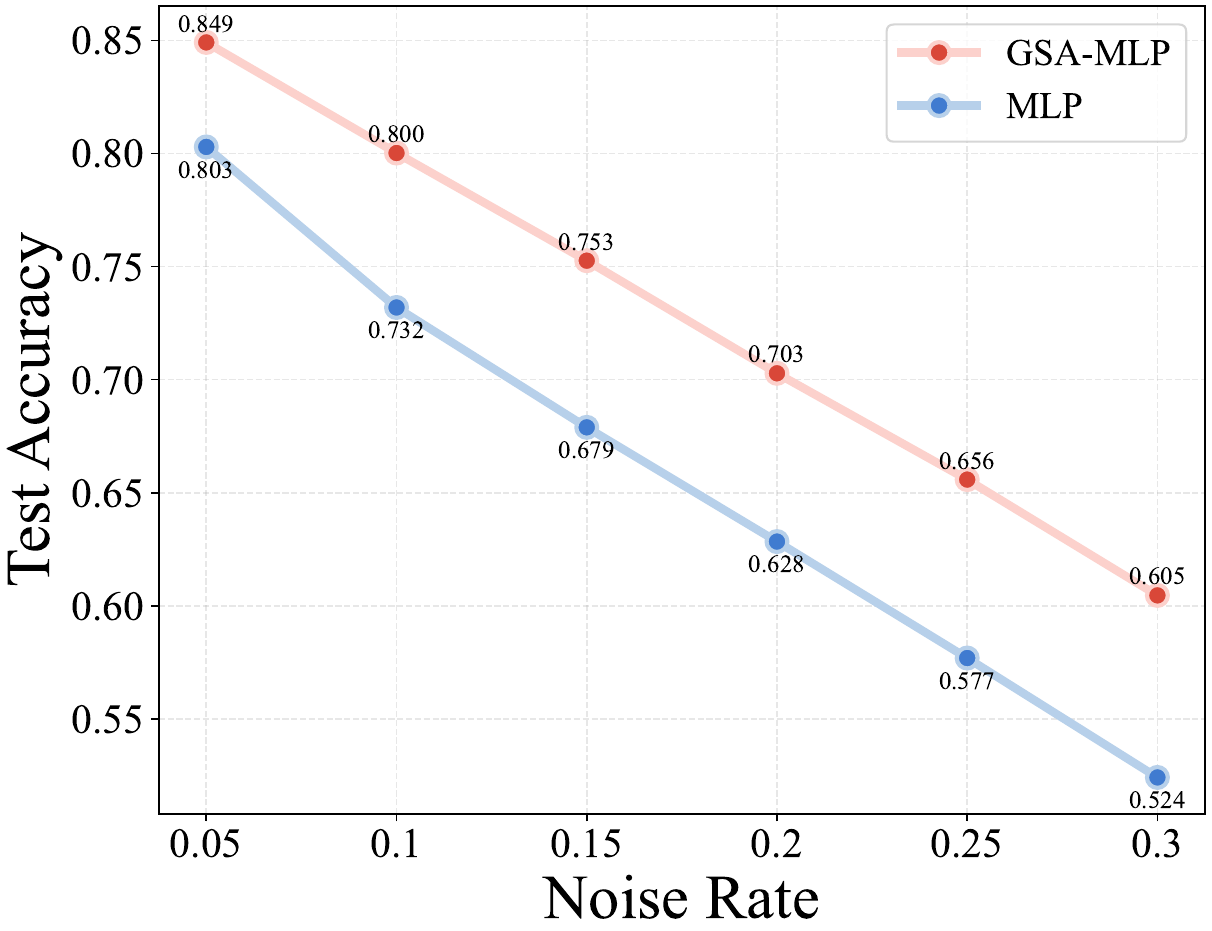}
    }
    \subfigure[MNIST (S15)]{
        \includegraphics[width=0.4\textwidth]{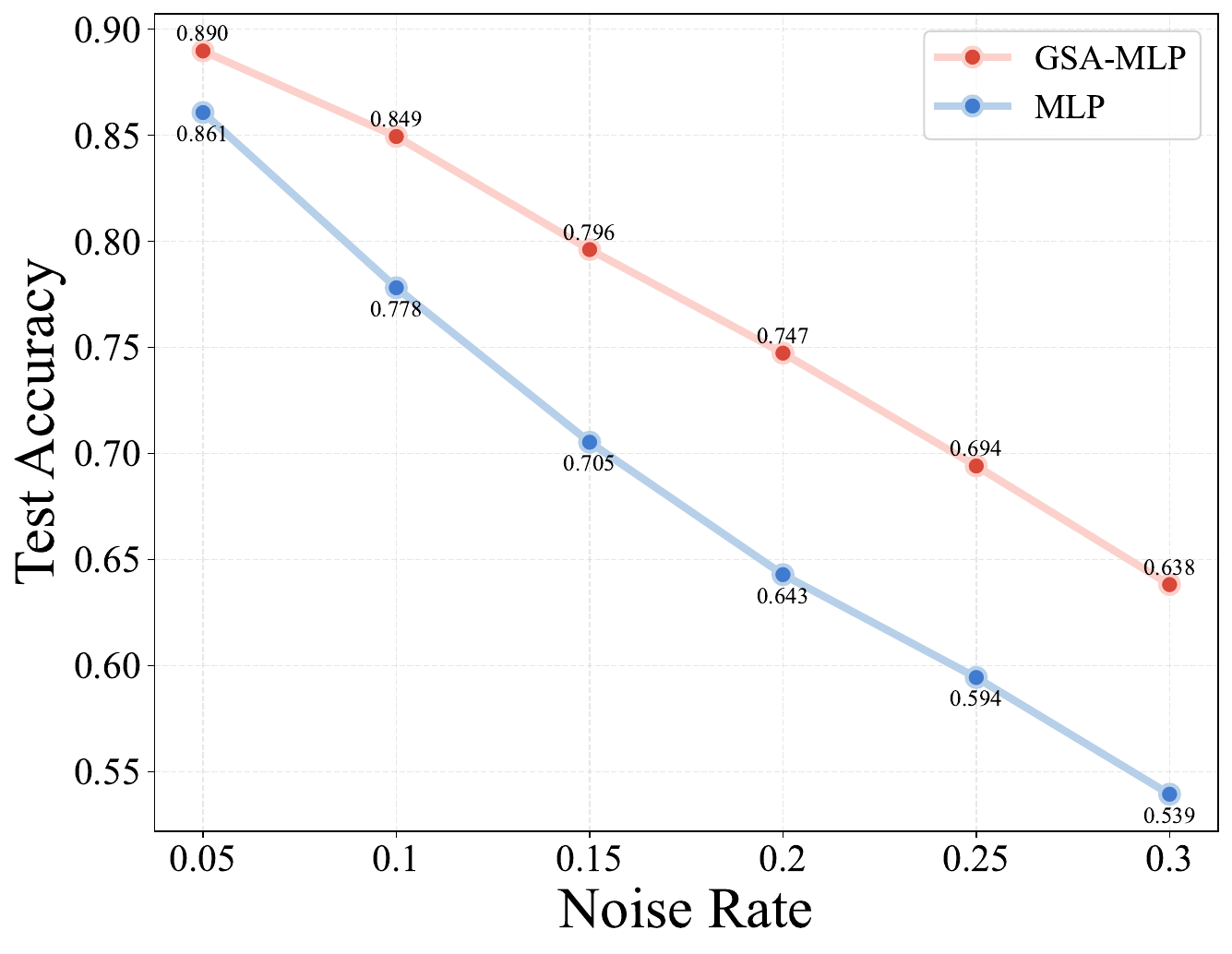}
    }
    \subfigure[organmnist\_sagittal (S14)]{
        \includegraphics[width=0.4\textwidth]{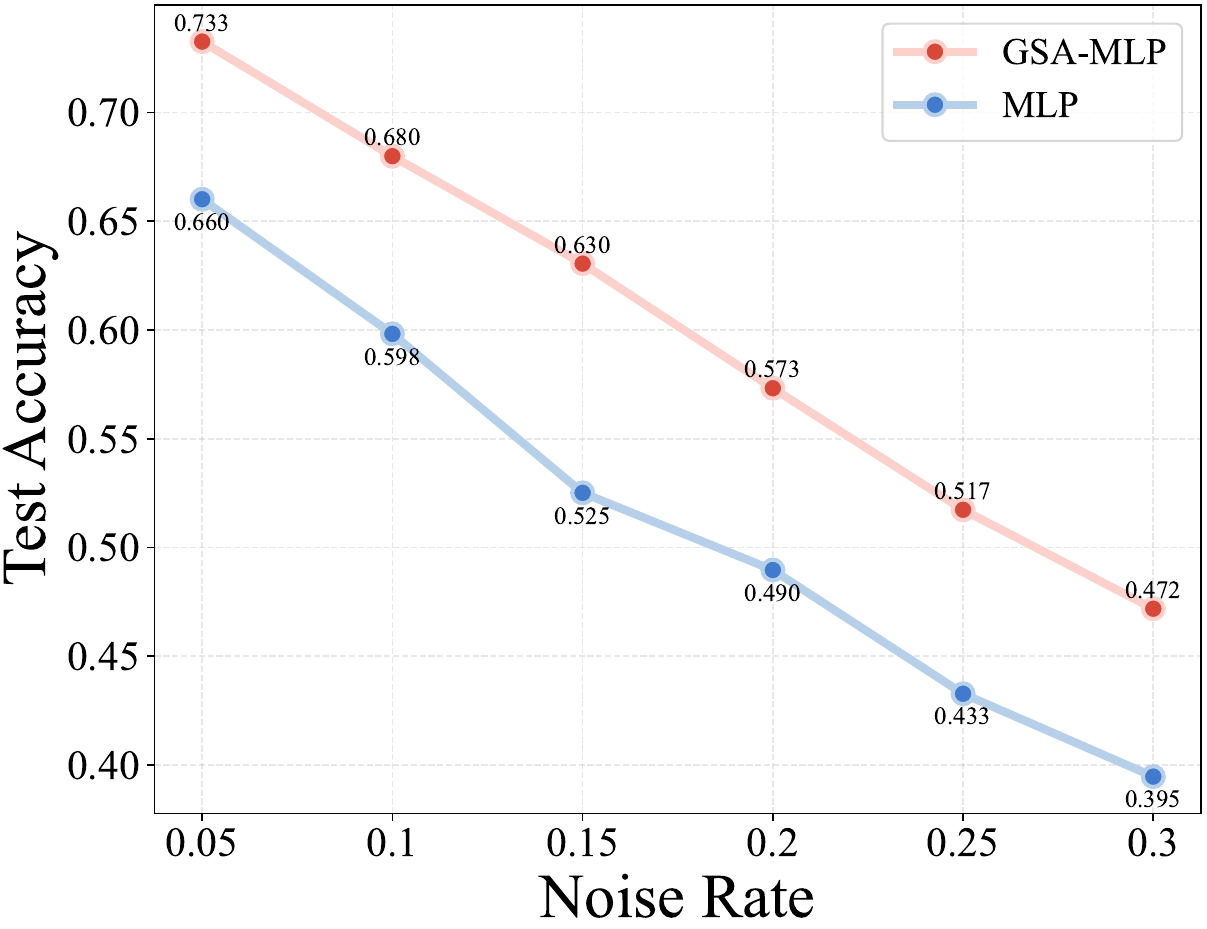}
    }
    \subfigure[USPS (S13)]{
        \includegraphics[width=0.4\textwidth]{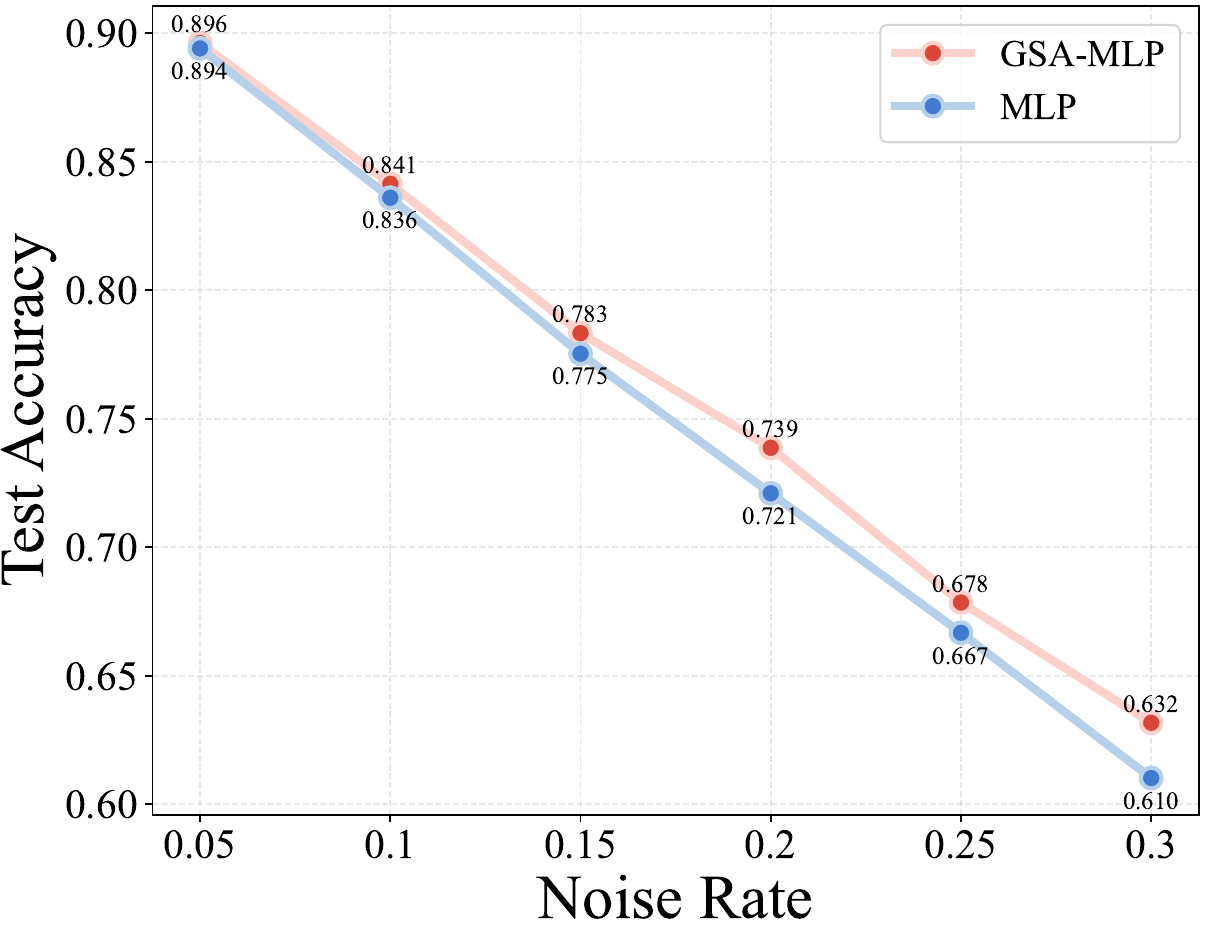}
    }
    \caption{Comparison of test $Accuracy$ on image datasets with various label noise rates.}
    \label{fig5}
\end{figure}

In real scenarios, image data often introduces label noise due to human errors in the labeling process, errors in automatic labeling tools, and other factors. Label noise presents a challenge to the training and prediction performance of classification algorithms. To evaluate the robustness of GAdaBoost.SA on image datasets under label noise scenarios, a comparative experiment is conducted, using MLP as the base classifier. The performance of GAdaBoost.SA is compared with that of a single MLP. Detailed results are shown in Figure \ref{fig5} and Table \ref{table7}, where GSA-MLP is used to denote GAdaBoost.SA with MLP as the base classifier.

As shown in Figure \ref{fig5}, at different noise levels, the test $Accuracy$ of GAdaBoost.SA on these image datasets is improved compared with that of a single MLP. This is because compared with GAdaBoost.SA, a single MLP lacks a processing mechanism for noise samples. Consequently, it is prone to overfitting or underfitting at the same noise level, resulting in performance degradation. In addition, GAdaBoost.SA performs more significantly on datasets with higher image quality, such as MNIST, than on datasets with lower image quality, such as USPS. The reason is that datasets such as MNIST have higher sample quality, allowing the incremental sampling strategy of GAdaBoost.SA to focus more accurately on key areas. Specifically, the GBs effectively capture the geometric distribution characteristics of the samples, particularly those close to the class boundaries. By focusing on the GBs containing misclassified samples during the iteration process, GAdaBoost.SA gradually strengthens the ability to distinguish easily confused samples, thereby improving the overall classification performance.

\begin{table}[H]
	\caption{Comparison of test $F1-score$ on image datasets with various label noise rates.}
	\label{table7}
	\centering
	\renewcommand\tabcolsep{1.5pt}
        \footnotesize
        \resizebox{\textwidth}{!}{
        \begin{tabular}{ccccccccccccc}
	\toprule[1pt]
\multirow{2}{*}{Dataset} &\multicolumn{2}{c}{Noise rate:5$\%$} & \multicolumn{2}{c}{Noise rate:10$\%$} &\multicolumn{2}{c}{Noise rate:15$\%$} &\multicolumn{2}{c}{Noise rate:20$\%$} & \multicolumn{2}{c}{Noise rate:25$\%$} &\multicolumn{2}{c}{Noise rate:30$\%$}
\\ \cmidrule(lr){2-3} \cmidrule(lr){4-5} \cmidrule(lr){6-7} \cmidrule(lr){8-9} \cmidrule(lr){10-11} \cmidrule(lr){12-13}
	 &GSA-MLP &MLP  &GSA-MLP &MLP &GSA-MLP &MLP &GSA-MLP &MLP &GSA-MLP &MLP &GSA-MLP &MLP
        \\ \hline
        S13   &\textbf{0.8901} &0.8872 &\textbf{0.8322} &0.8277 &\textbf{0.7729} &0.7642 &\textbf{0.7277} &0.7095 &\textbf{0.6681} &0.6530 &\textbf{0.6217} &0.5989      \\
        S14   &\textbf{0.6982} &0.6187 &\textbf{0.6440} &0.5618 &\textbf{0.5946} &0.4967 &\textbf{0.5356} &0.4533 &\textbf{0.4796} &0.3911 &\textbf{0.4366} &0.3646     \\
        S15   &\textbf{0.8890} &0.8599 &\textbf{0.8490} &0.7772 &\textbf{0.7954} &0.7040 &\textbf{0.7466} &0.6412 &\textbf{0.6933} &0.5921 &\textbf{0.6044} &0.5375     \\
        S16   &\textbf{0.8490} &0.8038 &\textbf{0.8002} &0.7336 &\textbf{0.7524} &0.6794 &\textbf{0.7025} &0.6286 &\textbf{0.6560} &0.5778 &\textbf{0.6373} &0.5244     \\
\toprule[1pt]
\end{tabular}
}
\end{table}

Similarly, as reported in Table \ref{table7}, GAdaBoost.SA consistently outperforms MLP in terms of $F1-score$ across all noise rates on these image datasets. This indicates that GAdaBoost.SA not only improves overall classification $Accuracy$ but also enhances the balance between precision and recall, which is particularly crucial for handwritten digit recognition under label noise. In such scenarios, certain digits (e.g., ‘3’ and ‘8’) are more prone to mislabeling due to their visual similarity, leading to an increased risk of misclassification in traditional classifiers like MLP. The superior $F1-score$ of GAdaBoost.SA suggests that its noise mechanism effectively mitigates such errors by refining decision boundaries and reducing the impact of mislabeled instances. In addition, the improvement in $F1-score$ is more pronounced on high-quality datasets such as MNIST, compared to lower-quality datasets like USPS. This aligns with the previous $Accuracy$ analysis.

\subsection{Ablation Study}
\label{Ablation}

\begin{figure}[H]
    \centering
    \subfigure[Noise rate: 5$\%$]{
        \includegraphics[width=0.7\textwidth]{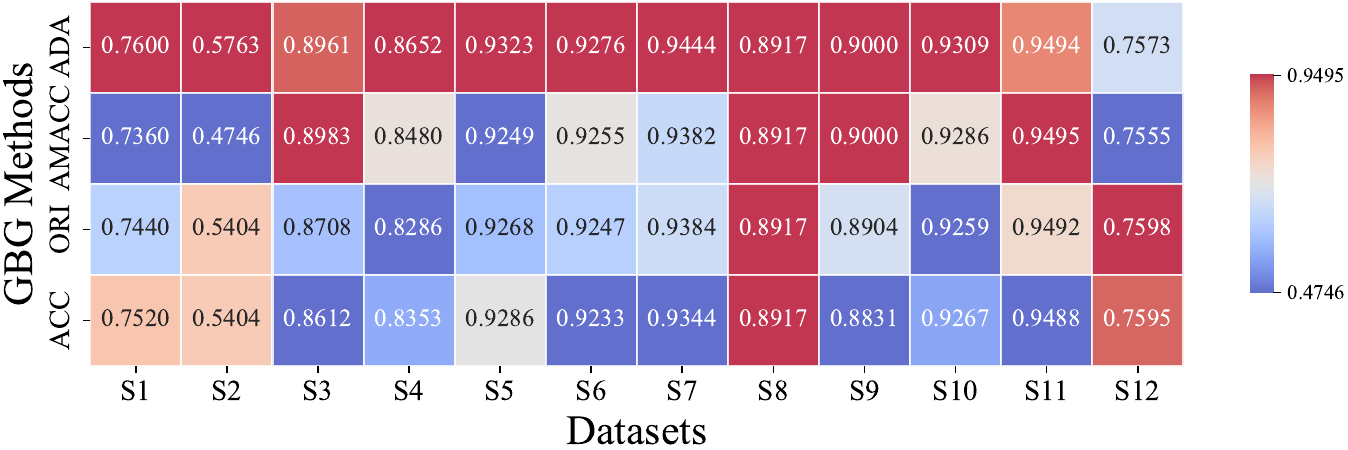}
    }
    \subfigure[Noise rate: 10$\%$]{
        \includegraphics[width=0.7\textwidth]{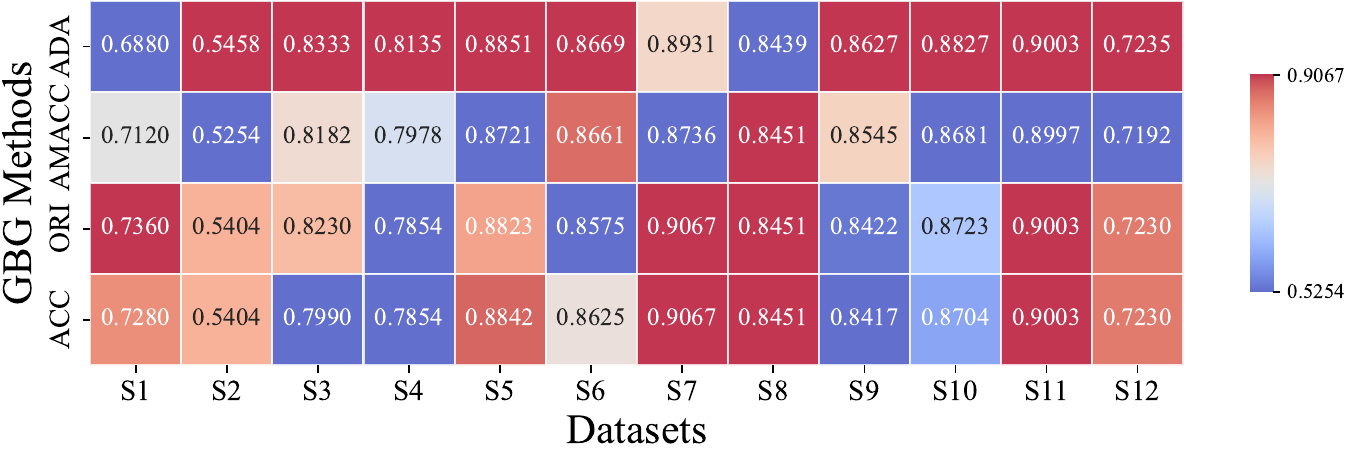}
    }
    \caption{Comparison of the effects of different GBG methods on test $Accuracy$ with $5\%$ and $10\%$ label noise rates.}
    \label{fig6}
\end{figure}

In this section, the effectiveness of the proposed POJG-Ens is verified. Figure \ref{fig6} shows the test $Accuracy$ of the proposed GBG-Ens and the existing POJG-based GBG methods, namely Ori-GBG, ACC-GBG, and AMACC-GBG, introduced in Section \ref{Granulation}, on 12 KEEL datasets at label noise rates of 5$\%$ and 10$\%$, respectively. These methods are used in the first stage of GAdaBoost.SA.

As shown in Figure \ref{fig6}, GBG-Ens based GAdaBoost.SA achieves the best test $Accuracy$ in most cases. The reason is that compared with POJG, POJG-Ens allows the GBG stage to capture more diverse information structures by requiring diversity. This, in turn, supports the second stage of GAdaBoost.SA to train good and diverse base classifiers, thereby effectively improving the performance of strong classifiers on complex data distributions.

\section{Conclusion}
\label{Conclusion}
This paper introduces GAdaBoost, a novel granular adaptive boosting framework designed to address the challenges of label noise and computational inefficiency in traditional AdaBoost. GAdaBoost consists of a two-stage learning process, where a data granulation stage is followed by an adaptive boosting stage that focuses on IGs instead of individual samples. Its implicit weighting mechanism makes GAdaBoost naturally base learner-agnostic, enabling seamless integration with various classifiers. To demonstrate its effectiveness and flexibility, an extension of SAMME, GAdaBoost.SA is proposed, integrating a GBG method and a GB-based SAMME algorithm. GAdaBoost.SA compresses data, preserves diversity, and mitigates label noise, leading to improved robustness and efficiency. By focusing on critical boundary samples while suppressing the influence of redundant or noisy samples, GAdaBoost.SA significantly enhances the performance of AdaBoost and its variants.

Experimental results on noisy datasets show that GAdaBoost.SA consistently achieves superior robustness when combined with diverse base classifiers such as CART, MLP, and SVM, while its efficiency is verified through experiments based on CART. These results highlight GAdaBoost's potential as a general-purpose boosting framework applicable to various classifiers and noisy environments. Future work will further explore its extension to other ensemble learning paradigms and large-scale real-world datasets.

\section*{Acknowledgement}
This work was supported by XXX.

\bibliography{mybibfile}

\end{document}